\documentclass[pdflatex,sn-vancouver-num]{sn-jnl}

\usepackage{graphicx}%
\usepackage{multirow}%
\usepackage{amsmath,amssymb,amsfonts}%
\usepackage{amsthm}%
\usepackage{mathrsfs}%
\usepackage[title]{appendix}%
\usepackage{xcolor}%
\usepackage{textcomp}%
\usepackage{manyfoot}%
\usepackage{booktabs}%
\usepackage{algorithm}%
\usepackage{algorithmicx}%
\usepackage{algpseudocode}%
\usepackage{listings}%

\usepackage{siunitx}
\usepackage{float}
\usepackage{caption}
\usepackage{etoolbox}
\usepackage{cleveref}
\usepackage{subcaption}

\usepackage[utf8]{inputenc}

\newcommand*{\ChristBERT}{Christ{BERT}}
\newcommand*{\ff}{F\textsubscript{1}}

\graphicspath{ {./figures/} }

\theoremstyle{thmstyleone}%
\theoremstyle{thmstyletwo}%

\theoremstyle{thmstylethree}%

\begin{document}

\title[The Word and the Way: Strategies for Domain-Specific BERT Pre-training in German Medical NLP]{The Word and the Way: Strategies for Domain-Specific BERT Pre-Training in German Medical NLP}




\author[1]{\fnm{Henry} \sur{He}}\email{henry.he@tum.de}
\equalcont{These authors contributed equally to this work.}

\author[2]{\fnm{Johann} \sur{Frei}}\email{johann.frei@informatik.uni-augsburg.de}

\author*[1,3]{\fnm{Raphael} \sur{Schmitt}}\email{raphael.schmitt@uniklinik-freiburg.de.de}
\equalcont{These authors contributed equally to this work.}

\affil[1]{\orgdiv{School of Computation, Information and Technology}, \orgname{Technical University of Munich}, \orgaddress{\city{Munich}, \country{Germany}}}

\affil[2]{\orgdiv{Chair of IT Infrastructure for Translational Medical Research}, \orgname{Faculty of Applied Computer Science, University of Augsburg}, \orgaddress{\city{Augsburg}, \country{Germany}}}

\affil[3]{\orgdiv{Institute of General Practice}, \orgname{Faculty of Medicine and Medical Center, University of Freiburg}, \orgaddress{\country{Germany}}}

\abstract{
\textbf{Background:} Digital healthcare generates vast amounts of clinical texts that hold potential for AI-assisted applications. However, existing German biomedical language models either rely on older architectures or are trained on limited data, which may hinder their performance in real-world settings.

\textbf{Methods:} To explore the impact of domain adaptation strategies in German clinical NLP, we developed a family of domain-specific RoBERTa-based language models, collectively referred to as \textit{ChristBERT} (\textbf{C}linical- and \textbf{H}ealthcare-\textbf{R}elated \textbf{I}ssues and \textbf{S}ubjects \textbf{T}uned BERT). To address the lack of large-scale German clinical corpora, we curated a 13.5\,GB dataset consisting of scientific publications, clinical texts, and health-related web content. Additionally, we employed data augmentation via translation of English clinical corpora. Three domain adaptation strategies were explored: continued pre-training, pre-training from scratch, and pre-training with domain-specific vocabulary adaptation.

\textbf{Results:} The resulting models were evaluated on three medical named entity recognition and two text classification tasks. Our models consistently outperformed four existing general-purpose and medical German models on four out of five tasks. The results demonstrate that the choice of domain adaptation strategy significantly influences downstream task performance. Based on the empirical results, pre-training from scratch is effective for highly specialized clinical texts, whereas continued pre-training is suited for more commonly written medical texts.

\textbf{Conclusions:} ChristBERT establishes a new state-of-the-art for German clinical language modeling. Our findings indicate that the optimal domain adaptation strategy is task-dependent and remains crucial, as adapted models consistently outperformed general-purpose
language models in our experiments. To support further research and application in German medical NLP, all developed models are publicly released.
}

\keywords{Natural Language Processing, Medical Informatics, Machine Learning, Electronic Health Records, Named Entity Recognition, Text Classification, Language Models, Biomedical Text Mining, Germany}



\maketitle

\section{Introduction}

The digitization of health services and clinical processes has resulted in the
healthcare industry generating an ever-increasing amount of textual data,
encompassing electronic health records, clinical notes, medical reports, and
discharge letters among many others. While structured data is frequently used
for health economics and registries, the aforementioned unstructured clinical
narratives are preferred by physicians to record patients' clinical information
due to their flexibility and efficiency, and make up to 40\% of the data
generated in current hospital systems~\cite{wang2018clinical,
dalianis2009stockholm}. The substantial potential of narrative text data to
support clinical applications was recognized early~\cite{sager1994natural,
borst1991textinfo,friedman1995architectural} and more recently, research efforts
have been directed towards developing medical applications assisted by
artificial intelligence (AI). Prominent applications include decision support
systems that assist healthcare professionals in their tasks, alleviating their
workload and providing better treatments for patients~\cite{zhou2022natural}.

However, the unstructured nature of textual data and the intricacies of the
biomedical field pose significant challenges for leveraging its potential. In
such a context, natural language processing (NLP) methods could structure that
information to support downstream clinical applications. Recent advancements in
NLP brought about by large-scale pre-trained language models based on the
Transformer~\cite{vaswani2017attention} architecture, introduced new ways
for extracting and analyzing the knowledge contained within the clinical texts.
Through extensive self-supervised training on vast corpora of text, a model can
acquire valuable representations of a language, producing highly effective
language models.

The success of Transformer-based models like BERT (Bidirectional Encoder
Representations from Transformers)~\cite{devlin2019bert} and its improved
version RoBERTa~\cite{liu2019roberta}, can be largely attributed to the use of
transfer learning expressed in the pretrain-finetune paradigm. In this paradigm,
a model initially goes through a resource-demanding training process, i.e.
\textit{pre-training}, using general-purpose textual data to learn the language
structure. This pre-training phase is self-supervised, eliminating the need for
labeled data by utilizing objectives like masked language
modeling~\cite{devlin2019bert}. The model is then \textit{fine-tuned} for
various tasks through a second, more cost-effective training round using a
smaller, labeled, and task-specific dataset that adjusts the model's weights to
fit the specific task and application domain at hand.

Direct application of general-purpose language models to a specific domain
might limit performance due to significant distributional differences
between general and target domains. Even within the same language,
domain-specific language can vary significantly from everyday language, leading
to the need of domain-specific models~\cite{arefeva2022tourbert}. This
particularly holds for the medical domain, where the language is highly
specialized and complex. Medical language features numerous acronyms that are
crucial for saving time and space, yet they can be ambiguous and require context
to be understood. Spelling errors are common, and there is an abundance of
abbreviations~\cite{tayefi2021challenges}. Moreover, the medical vocabulary is
highly specialized, as it is not typically used in everyday language, making it
unfamiliar to those outside the medical profession. When the target domain, such
as medicine, differs considerably from the pre-training data, models can be
improved by an additional phase of domain-adaptive training using large,
domain-specific corpora with the same pre-training objectives. 

Such specifically designed medical language models hold significant promise for
enhancing the efficiency and precision of medical document
handling~\cite{beltagy2019scibert, huang2019clinicalbert, peng2019transfer,
lee2020biobert}. For the German medical domain, the effectiveness of such models
has been demonstrated by BioGottBERT~\cite{lentzen2022critical} and
medBERT.de~\cite{bressem2024medbert}. However, the availability of open-source
biomedical corpora large enough for domain adaptation is limited, primarily due
to the sensitive nature of health-related data, and is largely confined to the
English language, given its established status as the language of science.
Despite these obstacles, advancing medical language models remains crucial, as
they have the potential to manage the large volumes of text produced in
hospitals every day.
In this work, we aimed to develop a new comprehensive German clinical language
model based on the RoBERTa architecture by building upon the foundation laid by
GeistBERT~\cite{scheibleschmitt2025geistbertbreathinglifegerman}, hereinafter referred to as \textit{\ChristBERT}: \textbf{C}linical-
and \textbf{H}ealthcare-\textbf{R}elated \textbf{I}ssues and \textbf{S}ubjects
\textbf{T}uned BERT. The main emphasis of this work lies in the construction of
a large German pre-training corpus, encompassing a diverse range of biomedical
and clinical texts. These sources provided a broad spectrum of medical
language data, fostering the model’s robustness and applicability. In order to
achieve this, we utilized a combination of mostly publicly available German
medical textual data and synthetic German domain texts by augmenting the corpus
with translated medical texts~\cite{edunov2018understanding}. This approach
involves translating a monolingual corpus using neural machine translation
models~\cite{ng2019facebook, costa2022no}, allowing us to leverage the vast
amount of public English medical texts available. Based on the constructed
corpus, we pre-trained \ChristBERT{} by using Whole Word Masking (WWM) and following three
different domain-adaptation strategies: (1) continued pre-training, (2)
pre-training from scratch with general-purpose vocabulary, and (3) pre-training from scratch with additional
prior vocabulary adaptation. In order to investigate the effects of the
different domain-adaptation approaches, we evaluated the performance of the
resulting models on two domain-specific downstream tasks: named entity
recognition and classification. The downstream task performance has been
thoroughly evaluated and compared to existing medical and general-purpose German
language models.

\section{Related Work}

Past developments in medical NLP research have seen the creation of mature
systems for extracting information from English clinical texts like
MetaMap~\cite{aronson2010overview}, cTAKES~\cite{savova2010mayo},
MedLEE~\cite{friedman1995architectural, friedman2000broad} and
CLAMP~\cite{soysal2018clamp}. These systems have been used for various tasks such as named entity
recognition (NER), relation extraction, and information retrieval. Additionally,
open competitions such as Informatics for Integrating Biology and the Bedside
(i2b2)~\cite{uzuner20112010}, National NLP Clinical Challenges
(n2c2)~\cite{henry20202018, stubbs2019cohort}, and CLEF eHealth
\cite{crestani2019experimental} challenge from the Conference and Labs of the
Evaluation Forum (CLEF) promote data and model sharing, further advancing the
medical NLP field. The systems developed to date encompass rule-based,
machine-learning-based, and hybrid models. While rule-based methods were
essential in early developments, the performance of these systems is limited by
their reliance on hand-crafted rules and lexicons, which are difficult to
maintain and generalize across different clinical settings. 

In order to overcome these challenges, current research emphasizes
machine-learning techniques. In particular, deep-learning approaches like
recurrent neural networks (RNN) and convolutional neural networks have been widely
used in recent years due to their ability to achieve superior performance with
adequate training data. Unlike traditional machine-learning methods, deep neural
networks typically use methods such as Word2Vec~\cite{mikolov2013distributed},
GloVe~\cite{peters2018dissecting}, or FastText~\cite{joulin2017bag} to represent
words as vectors. These methods create word embeddings by learning relationships
between words from large text corpora, eliminating the need for manual feature
engineering. Nevertheless, these methods represent all possible meanings of a
word in a single vector, making them unable to distinguish between different
word senses based on the surrounding context. Vaswani et al.
\cite{vaswani2017attention} introduced a new model able to provide
contextualized word representation called the Transformer. Originally designed
for neural machine translation, the Transformer addresses two limitations of
RNNs: lack of parallelization and handling of long-range dependencies. It relies
on the self-attention mechanism, which differentially weighs parts of the input.
Since it operates without recurrence, it is more parallelizable and
computationally efficient than RNNs.

In 2019, Devlin et al.~\cite{devlin2019bert} utilized parts of the original
Transformer architecture to develop BERT, achieving state-of-the-art results in
numerous NLP tasks. Performance of these large-scale language models heavily
depends on the underlying data used for pre-training. A homogeneous text corpus
generally leads to a poorer performing model compared to one trained on diverse
text corpora of high variance~\cite{martin2020camembert}. Initially, much of
BERT research was conducted with English texts, followed by efforts in
multilingual approaches~\cite{conneau2020unsupervised}. While multilingual
models were trained on extensive texts from numerous languages, it has been
shown that single language models outperform these and are even beneficial in
terms of efficiency, pre-training efforts, and downstream task performance as
they demand fewer computational resources and smaller datasets compared to the
extensive and diverse data required for multilingual
models~\cite{scheible2020gottbert, chan2020german, martin2020camembert}. In
particular, single-language models trained with the Open Super-large Crawled
ALMAnaCH coRpus (OSCAR)~\cite{suarez2019asynchronous} demonstrated strong
performance, benefiting from the corpus's size and variability. Notable examples
include CamemBERT~\cite{martin2020camembert} for French,
GottBERT~\cite{scheible2020gottbert} for German, and BERTje~\cite{de2019bertje}
for Dutch.

With the increasing use of Transformer-based models in NLP, there is a growing need in the clinical domain for language models that are not only accurate but also efficient, resource-conscious, and suitable for local processing. In settings with limited computational resources and strict data privacy requirements, small yet high-performing domain-specific models can provide substantial benefits. Continued pre-training on in-domain data has proven effective for enhancing performance on specialised clinical tasks. In the biomedical field, the pioneering and most recognized pre-trained model is
BioBERT~\cite{lee2020biobert}, which shares the same architecture as BERT.
Following a domain-adaptation strategy, BioBERT starts with BERT weights
pre-trained on general texts and then refines these weights using biomedical
corpora, surpassing the original model and achieved state-of-the-art
performance in numerous biomedical text mining tasks, such as clinical concept
recognition, gene-protein relation extraction, and biomedical question
answering. To gather sufficient open-source biomedical data, the authors
utilized repositories like PubMed~\cite{white2020pubmed} and PMC~\cite{pmcoa},
obtaining 4.5 billion words from abstracts and 13.5 billion words from full-text
articles. A similar method is employed by SciBERT~\cite{beltagy2019scibert},
which retains the original BERT configuration but substitutes the initial
general corpora with 1.14 million scientific articles randomly chosen from
Semantic Scholar. This dataset consists of 82\% broad biomedical domain papers
and 18\% computer science domain papers. By training from the ground up on
biomedical data, SciBERT can utilize a custom dictionary that better represents
the domain-specific word distribution. Med-BERT~\cite{liu2021med} is the first model fully trained on hospital data,
particularly semi-structured electronic health records, leading to enhanced
performance in subsequent prediction models. These approaches have since been
refined, either by updating the model architecture to use BERT variants or by
expanding the biomedical corpus with additional sources beyond scientific
literature~\cite{huang2019clinicalbert, peng2019transfer}.

The extensive range of biomedical and clinical BERT-based models benefit from
the abundance of publicly available biomedical data in English, such as
MIMIC~\cite{johnson2023mimic,johnson2023mimicnote}, the largest open-access
dataset of medical records, and extensive repositories of biomedical scientific
literature~\cite{white2020pubmed}. However, most other languages lack access to
these valuable resources, making it challenging to achieve the same level of
performance as their English counterparts. Despite this, researchers from
various countries have endeavored to pre-train non-English biomedical models,
utilizing local and often non-public biomedical text collections. They have
either trained new models from scratch~\cite{akhtyamova2020named} or applied
biomedical domain adaptation to multilingual~\cite{rubel2020biobertpt} or
monolingual~\cite{copara2020contextualized} versions of BERT.

For what concerns the German language, advancements in medical language models
are significantly delayed and are often propelled solely by commercial software
or localized applications~\cite{starlinger2017improve}. Stringent data
protection laws impede data sharing, leading clinics to restrict data usage to
internal purposes~\cite{hellrich2015sharing}. These obstacles hinder the sharing
of datasets and models, as well as the organization of open challenges involving
German datasets. In spite of these challenges, there have been notable
initiatives in recent years: Datasets such as JSynCC~\cite{lohr2018sharing} and
GGPONC~\cite{borchert2022ggponc} have been released, containing German
biomedical language texts that are not subject to data protection concerns.
Recently, the introduction of the BRONCO150~\cite{kittner2021bronco150} corpus,
which includes de-identified discharge letters, and
GPTNERMED~\cite{frei2023gptnermed}, which leverages large language models, has
further expanded the availability of German medical text data. Additionally, the
CLEF eHealth challenge in 2019 provided a dataset of non-technical summaries of
animal studies to be classified according to the International Classification of
Diseases and Related Health Problems (ICD-10)~\cite{clef2019nts, clef2019test,
world1992icd}. A study by~\cite{sanger2019classifying} utilized the multilingual
BERT version (mBERT) to classify these summaries, demonstrating that mBERT
significantly outperformed a baseline Support Vector Machine model. To
incorporate advances in general German language models,
\cite{lentzen2022critical} introduced BioGottBERT, a model pre-trained on open
medical German texts from Wikipedia and scientific abstracts, which demonstrated
superior performance over its generalized counterpart GottBERT on medical tasks.
Subsequently, the authors of~\cite{bressem2024medbert} proposed medBERT.de, in
order to address the limited training data size and narrow scope on merely one
medical subarea by using 3.8 million radiology reports, achieving promising
results in classification tasks. While BioGottBERT was trained on a relatively
small corpus slightly less than 1 GB of text, medBERT.de significantly expanded
its training corpus to 10 GB, incorporating a wider variety of sources. However,
its BERT architecture has been improved by its optimized version RoBERTa as
recently demonstrated for German by the GeistBERT model \cite{scheibleschmitt2025geistbertbreathinglifegerman}. GeistBERT
reiterated on GottBERT~\cite{scheible2020gottbert}, by using Whole Word Masking
(WWM) and continued pre-training on a significantly more varied and larger
general-domain corpus, thereby establishing state-of-the-art performance on
various German NLP benchmarks.

\section{Methodology} \label{chap:methodology}


\subsection{Corpus Creation} \label{sec:corpus}
Main shortcomings of existing German medical domain models include the limited
availability of training data due to the sensitive nature of medical information
and strict data privacy regulations. Furthermore, many existing biomedical
Transformer models~\cite{lentzen2022critical, bressem2024medbert} are
pre-trained or evaluated on proprietary datasets, hindering independent model
verification and validation. Previous studies~\cite{martin2020camembert,
dada2023impact, bressem2024medbert} concluded that training data diversity and
quantity are more important than excessive data cleaning, which insignificantly
affected downstream performance. Following these findings, we compiled a 13.5 GB
large and highly varying German biomedical and clinical corpus, focusing on data
quantity over quality. In order to mitigate the aforementioned shortcomings, we
primarily relied on public datasets with only two private data sources included,
to foster transparency and accessibility of the \ChristBERT{} models.
Tab.~\ref{tab:corpus_stats} summarizes the pre-training corpus, including
descriptive statistics about the number of documents, sentences, words, and size
of each incorporated dataset.

\begin{table}[htb]
    \centering
    \begin{tabular}{lrrrr}
    \toprule
    \bfseries Dataset & \bfseries \# Documents & \bfseries \# Sentences & \bfseries \# Words & \bfseries Size (MB) \\
    \midrule
    Hpsmedia & 277,357 & 16,314,452 & 405,316,578 & 3,117 \\
    Springer Nature & 258,000 & 14,158,151 & 259,284,884 & 1,984 \\
    PubMed Central & 90,273 & 8,644,017 & 220,033,966 & 1,609 \\
    PhD Theses & 7,486 & 4,665,850 & 90,380,880 & 646 \\
    Medical Wikipedia & 75,585 & 3,254,135 & 49,594,111 & 362 \\
    MIMIC-IV Notes & 330,486 & 49,351,295 & 733,952,748 & 5,310 \\
    Web Crawl & 93,642 & 4,024,816 & 68,797,358 & 512 \\
    \midrule
    \bfseries Summary & \bfseries 1,132,829 & \bfseries 100,412,716 & \bfseries 1,827,360,525 & \bfseries 13,540 \\
    \bottomrule
    \end{tabular}
    
    \caption[Overview of datasets contained in the pre-training corpus]{
        Overview of datasets contained in the pre-training corpus. The table
        provides details about each dataset, including the number of documents,
        sentences, words, and their size in megabytes. The final corpus includes
        all listed datasets and amounts to roughly 13.5 GB of pre-training
        data.}
    \label{tab:corpus_stats}
\end{table}


\subsubsection{Hpsmedia}
Hpsmedia is a German publisher specializing in medical content primarily
targeted at healthcare professionals. Hpsmedia publishes three healthcare
journals \textit{Pflegewissenschaften (Nursing Sciences)}, \textit{Pädagogik der
Gesundheitsberufe (Pedagogy of Health Professions)} and \textit{Geschichte der
Gesundheitsberufe (History of Health Professions)}, which are available in print
and online. All journals publish articles in German and are peer-reviewed by
experts in the respective fields according to the international reviewing
standard BMJ \cite{smith2006peer}. The articles cover a wide range of topics
within the healthcare domain including aspects of health and nursing care,
pedagogy, didactics, curricula, education in healthcare professions and the
history of healthcare professions. We were kindly provided with the full-text
content of the journals in CSV format by Hpsmedia. The CSV files were processed
using the \textsc{Pandas} \cite{mckinney2010data} Python library to extract the
text content of the articles, which was then included in the pre-training
corpus. The Hpsmedia dataset consists of 277,357 documents totaling to 3,117 MB
of data.

\subsubsection{Springer Nature}
Springer Nature is a prominent global publisher of academic content, known for
its extensive collection of high-quality journals, books, and research materials
across various disciplines, including science, technology, and medicine.

For the extraction of text from Springer Nature publications, the Springer
Nature API~\cite{springernature} was utilized. The API offers multiple
endpoints, e.g. metadata, full-text (TDM) as well as a wide range of constraint
parameters to filter for desired publications, which are returned in XML format.
This allowed for a systematic filtering for open-access publications in German.
For our purposes, the open-access API was first queried for metadata of articles
and books related to the subjects of \textit{biomedicine}, \textit{public
health}, \textit{pharmacy}, \textit{dentistry} and \textit{life sciences}. The
returned XML data was then processed to extract abstracts and Digital Object
Identifiers (DOI) of each publication, respectively. Subsequently, the set of
DOIs was used to make bulk API calls to the TDM endpoint to subsequently fetch
the full-text content of the publications. In a final step, the extracted
abstracts and full-text content were both incorporated into the pre-training
corpus accounting for a total of 258,000 documents and 1,984 MB of data.

\subsubsection{PubMed Central}
\begin{figure}[!htbp]
    \centering
    \includegraphics[width=0.5\textwidth]{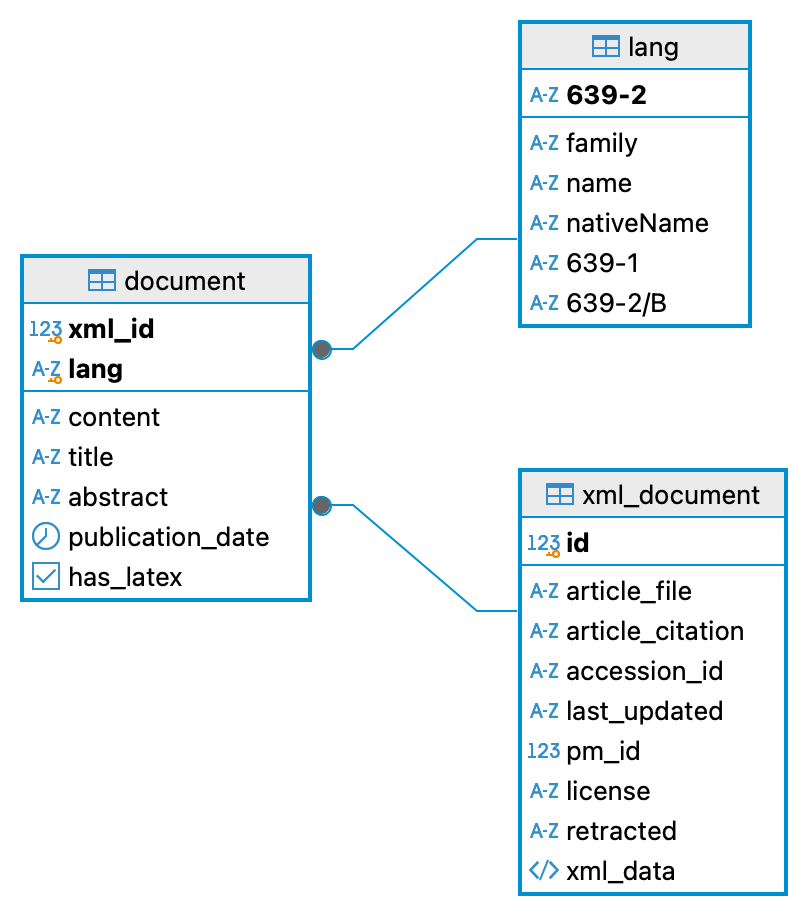}
    \caption[Database design for PMC translation]{The diagram shows the database
    relations for translation management. The raw XML files with their metadata
    are stored in the \texttt{xml\_document} table, while the \texttt{document}
    table contains the extracted text data with both PubMed ID and the language
    of a document as its primary keys. Languages are foreign keys to the
    corresponding entity in the \texttt{lang} relation according to the ISO
    639-3 standard.}
    \label{fig:pmc_erd}
  \end{figure}

PubMed Central (PMC) is a free digital repository of full-text scientific
literature in the field of biomedicine and life sciences and created as an
extension of PubMed \cite{white2020pubmed}, which holds bibliographic references
and abstracts for essentially all publications in the biomedical sciences. Both
repositories are maintained by the National Center for Biotechnology Information
(NCBI), a part of the United States National Library of Medicine (NLM). The PMC
archive provides access to a collection of over 10 million research articles,
reviews, and other scientific publications from a wide range of biomedical and
life science journals. Not all articles in PMC are available for text mining or
other reuse as many are under copyright. The \textit{PMC Open Access Subset}
\cite{pmcoa} contains those articles made truly freely available to the public
under Creative Commons or similar licenses that allow more liberal
redistribution and repurpose than the majority of licensed and copyrighted
articles from subscription access journals deposited in PMC.
PMC stores content in XML format, which is structured according to the Journal
Article Tag Suite (JATS) standard, a widely used archival markup format for
journal articles. The JATS XML files are made available by NLM for bulk download
through their PMC FTP Service \cite{pmcoa}. We downloaded the December 2024
baseline package of the PMC Open Access Subset and transferred the XML files
with appropriate metadata such as PubMed ID and publication date to a
\textsc{PostgreSQL} database for further processing. The database design is
shown in Fig.~\ref{fig:pmc_erd}, which is represented by an entity-relationship
diagram. The XML files and their corresponding metadata are stored in the
\texttt{xml\_document} table by leveraging the native support of
\textsc{PostgreSQL} for XML data types. For our needs, we extracted the title,
abstract, full-text content and language of the articles from the XML markup by
utilizing the \textsc{Pubmed Parser} \cite{achakulvisut2020} Python library,
which supports parsing of the JATS XML format. The extracted text data was then
stored in the \texttt{document} table, which contains the PubMed ID and language
of each document as its primary keys. The language of each document is
represented as a foreign key to the \texttt{lang} table, which contains the ISO
639-3 language codes. The \texttt{lang} table is used to ensure data integrity
and consistency across the database.
  
\begin{figure}[htbp]
    \centering
    \includegraphics[width=0.9\textwidth]{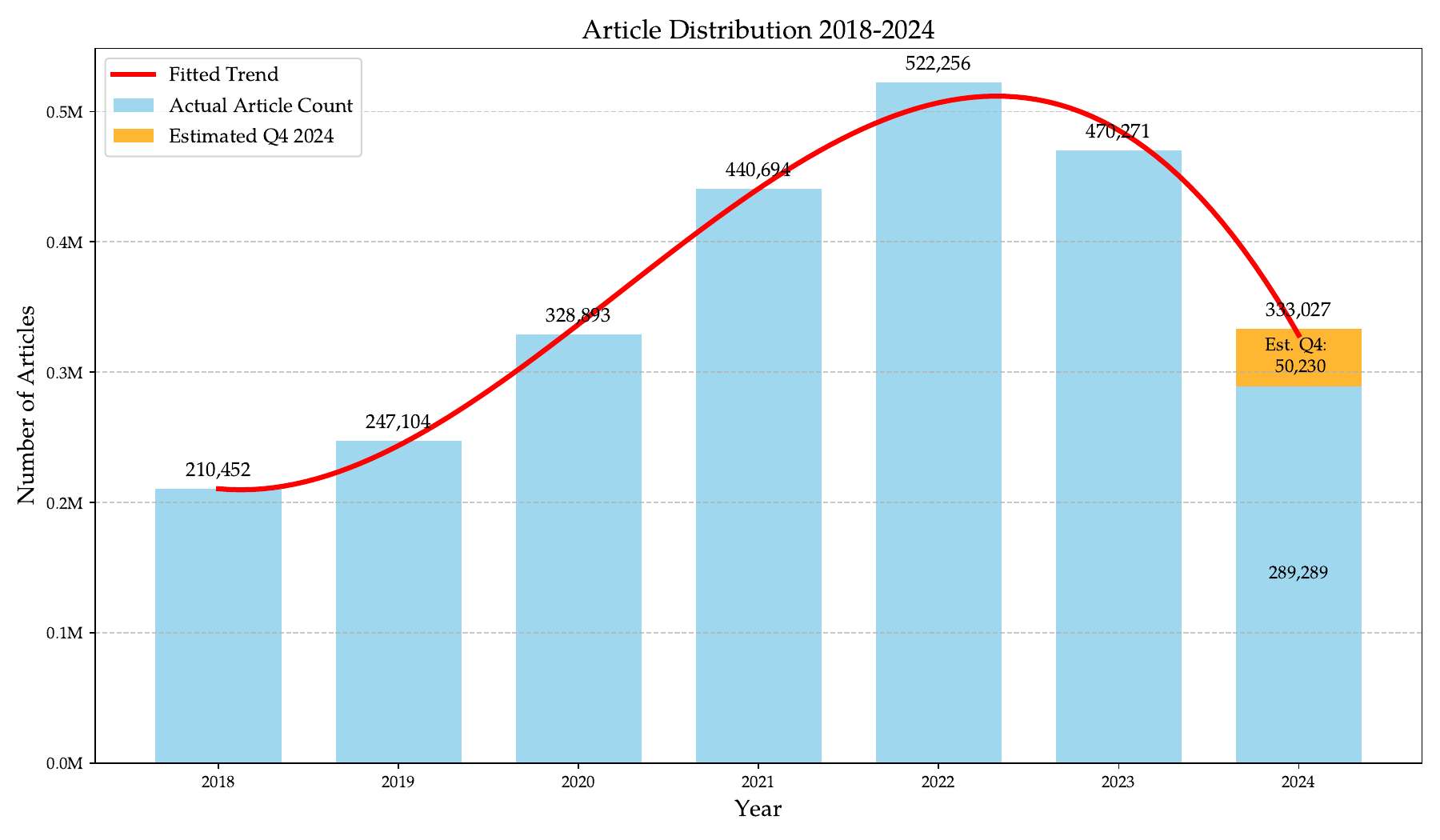}
    \caption[Distribution of PMC articles from 2018-2024]{This histogram shows
    the annual number of articles published in PMC from 2018-2024, with a fitted
    trend line indicating the overall growth and decline of article counts. The estimated article count for Q4 2024 was approximated based on the maximum Q4 article count observed in previous years (50,230).}
    \label{fig:histogram}
\end{figure}

\begin{figure}[htbp]
    \centering
    \includegraphics[width=0.9\textwidth]{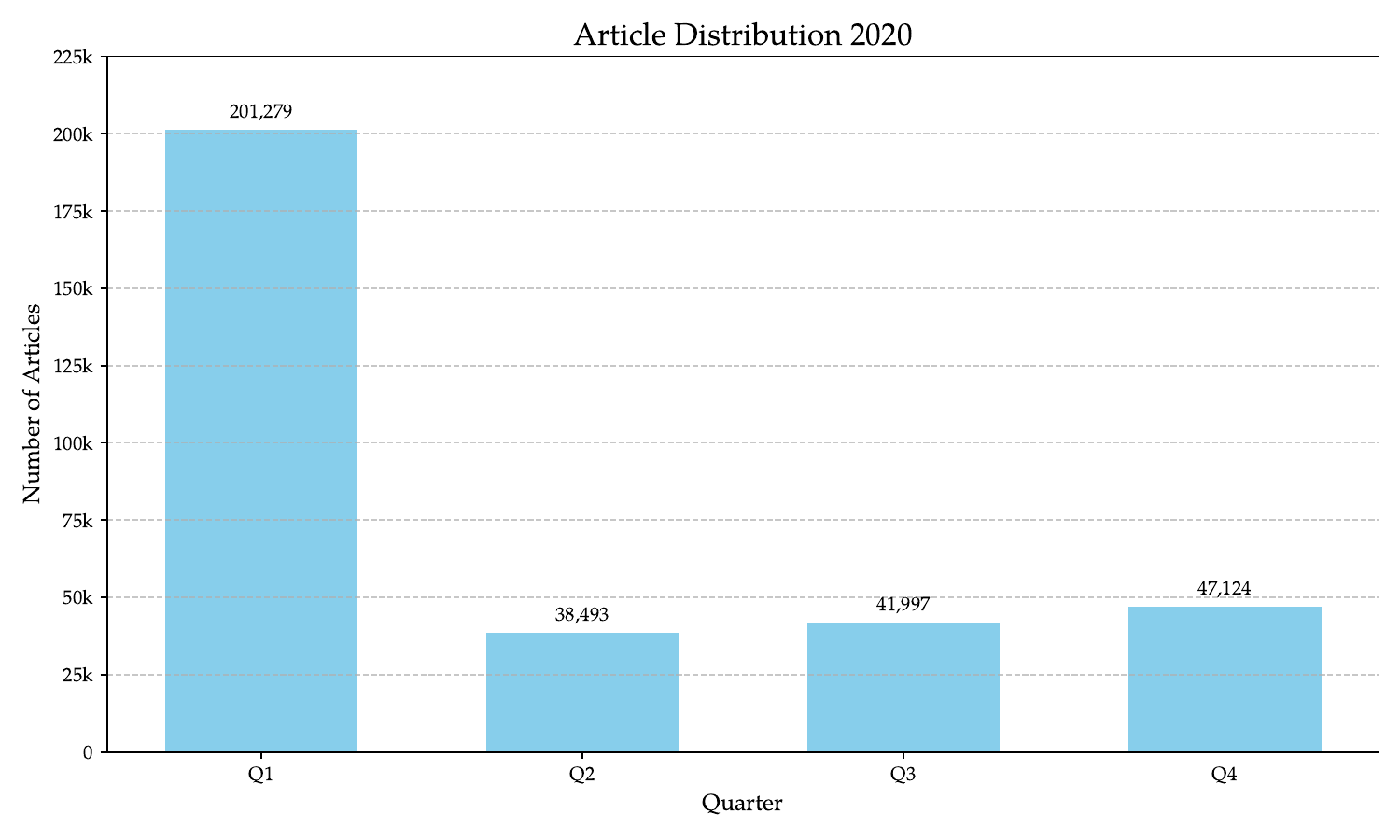}
    \caption[Quarterly distribution of PMC articles in 2020]{The histogram
    shows the quarterly number of articles published in PMC in 2020, with a
    peak in Q1 at 201,279 articles, followed by a drop in the subsequent
    quarters.}
    \label{fig:histogram_2020}
\end{figure}

\begin{figure}[H]
    \centering
    \includegraphics[width=0.5\textwidth]{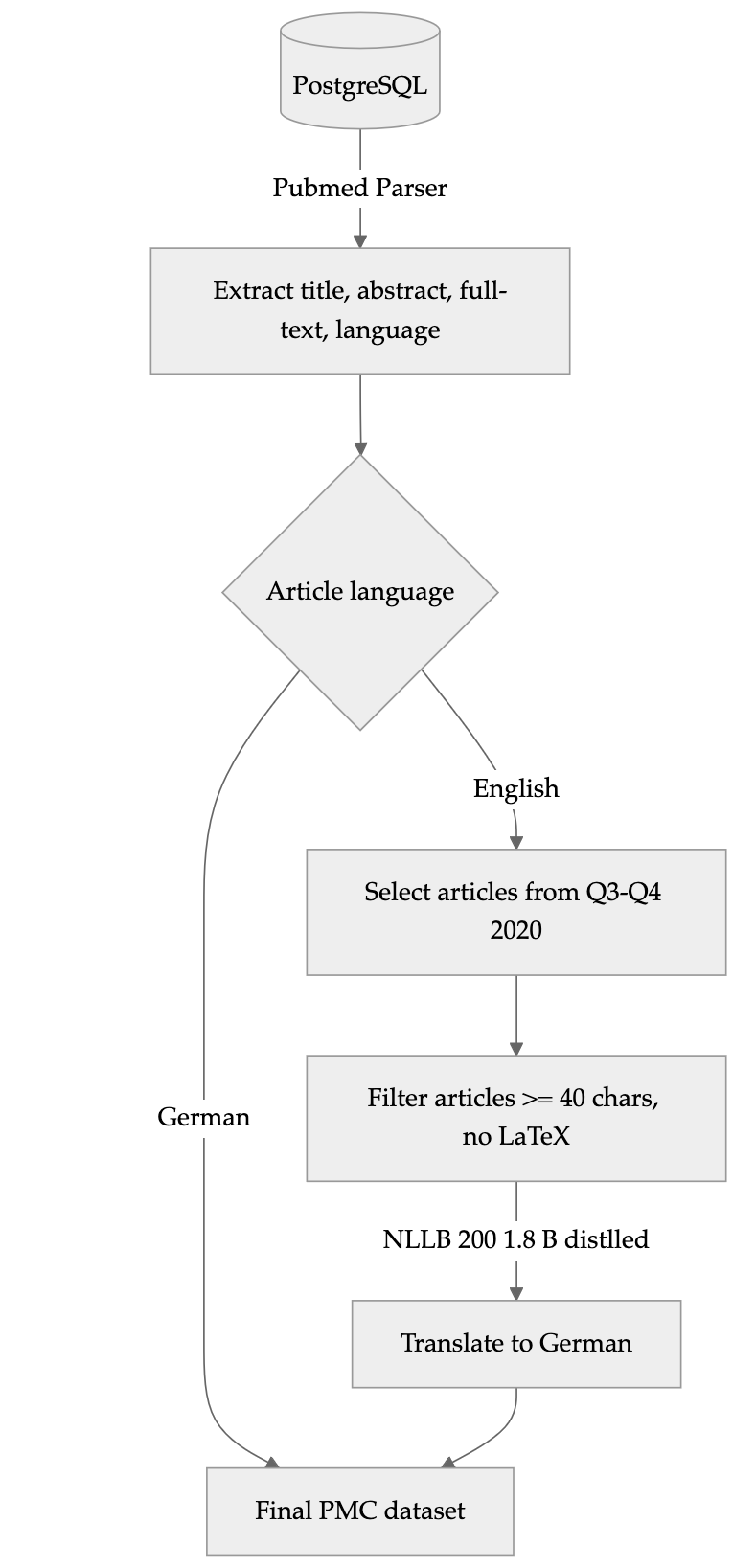}
    \caption[Flowchart of the PMC translation process]{This flowchart
    illustrates the sequential steps involved in the translation process of
    articles from PubMed Central (PMC). It begins with extracting article
    details from \textsc{PostgreSQL} using \textsc{Pubmed Parser}, followed by
    language-based filtering to select English articles from Q3 and Q4 of 2020.
    Articles are then filtered based on length and \LaTeX{} content, followed by
    translation to German using the NLLB 200 1.3B distilled model.}
    \label{fig:translation}
\end{figure}

In order to leverage the large amount of English-language content available in
PMC, we translated the English articles to German using the \textsc{NLLB
200}~\cite{costa2022no} neural machine translation model in its 1.3 billion
distilled variant. Translation was performed on two Nvidia GeForce RTX 3090 24
GB GPUs, while leveraging the \textit{NLLB-API} \cite{nllbAPI} library for
parallel processing. The translation posed a significant computational
challenge, which was addressed by limiting the publications to be translated to
those published in the third and fourth quarters of 2020. This decision was
based on an analysis of article distribution over the past seven years, which is
depicted in Fig.~\ref{fig:histogram}. The analysis revealed a notable peak in
publications in 2022, potentially influenced by the COVID-19 pandemic and the
emergence of generative AI. To ensure that the translated content was not overly
biased towards the COVID-19 pandemic and mitigate the presumably uniform writing
style resulting from generative AI, we selected the year 2020 for our
translations. Likewise, given our computational constraints, we chose the third
and fourth quarters of 2020, as the quarterly distribution of articles in that
year indicated a more feasible volume of publications as seen in
Fig.~\ref{fig:histogram_2020}. Translated documents were saved back to the
database in the \texttt{document} table, but with an updated language key set to
\texttt{de} for German. Further data filtering encompassed the removal of
articles with less than 40 characters and those containing \LaTeX{} markup.
Fig.~\ref{fig:translation} summarizes the described steps for PMC translation as
a flowchart. The translated and natively German articles were then combined into
a single dataset, resulting in a total of 90,272 documents and 1,609 MB of text
data.


\subsubsection{PhD Theses} 
In this work, we also included a collection of 7,486 open-access German-language
dissertations and postdoctoral dissertations from Charité University Hospital,
Germany's largest university hospital. At the joint medical faculty of Humboldt
University and Free University of Berlin, electronically published documents,
doctoral and habilitation theses, as well as research data are made available to
the public through the university's institutional repository \textsc{Refubium}
\cite{refubium}. The documents were downloaded in bulk as PDF files and
subsequently converted to plain text. Data cleaning involved removing sentences
that lacked German stop words and excluding theses under 15 pages in length.
This process ensured the inclusion of only relevant, high-quality text data. In
total, 646 MB of text was extracted from the PhD theses.

\subsubsection{Medical Wikipedia}
Wikipedia curates entry pages on the encyclopedia about particular subject areas
in so-called \textit{portals}. Each portal acts as a hub, bringing together key
articles, images, and resources about the respective topic. Portals are
particularly useful for getting an organized overview or exploring related
subtopics without searching through individual articles. We utilized the German
Wikipedia portal on medicine in order to extend our pre-training corpus with
freely available texts on medical topics, which were collectively authored and
editorially proofread by a diverse community of volunteer contributors.
Wikipedia does not offer an API for bulk data retrieval, but instead provides an
export interface \cite{wikipediaexport} for downloading specified wiki pages in
a special XML format. These XML files follow a schema specific to
\textsc{MediaWiki}, the software behind Wikipedia, initially intended for
importing into another \textsc{MediaWiki} installation but also allows for
further processing and analysis.

The export interface expects either a list of page titles or a category name,
which it resolves into a list of pages related to the given category. Since our
objective is to crawl the entirety of the medical portal, we implemented a
breadth-first search algorithm on Wikipedia's export interface, employing
\textsc{Selenium WebDriver}~\cite{gojare2015analysis} to traverse the category
tree of the portal. The algorithm starts at the root category
\texttt{Portal:Medizin} and recursively visits each subcategory, collecting the
titles of all pages contained within. The page titles are then used to download
the corresponding XML files in bulk, creating a dump of the entire German
medical portal. 
The \textsc{MediaWiki} XML files are parsed using Python’s \texttt{ElementTree} module to extract page contents. Wikitext formatting is then removed using \texttt{MediaWiki Parser from Hell} \cite{kurtovic_earwigmwparserfromhell_2025}, resulting in clean plain text documents. The German Wikipedia portal on medicine contributes a total of 75,585 documents and 362 MB of data to the pre-training corpus.

\subsubsection{MIMIC-IV Notes}
Medical Information Mart for Intensive Care IV (MIMIC-IV)
\cite{johnson2023mimic} is a large and freely accessible electronic health
record dataset comprising various health-related data acquired during routine
clinical care of patients admitted to critical care units of the Beth Israel
Deaconess Medical Center in Boston, MA, USA. MIMIC-IV constitutes the fourth
edition of the dataset, containing data of over a decade from 2008 to 2019 and
covering a wide range of information such as patient measurements, orders,
diagnoses, procedures, treatments, and clinical notes. 

For our corpus, we specifically chose to utilize the \textit{clinical notes}
\cite{johnson2023mimicnote} subset of the MIMIC-IV database as it is made up of
discharge summaries written in the form of free text, which is well suited for
training contextual language models. The 330,485 discharge summaries from
145,915 hospitalized patients are organized into sections including chief
complaint, history of present illness, past medical history, brief hospital
course, physical exams, and discharge diagnoses. These free-text notes were
acquired from the hospital system and de-identified by the authors using a
combined automatic approach of custom rules and a neural network trained on
de-identification, cast as a NER task. 

The note subset of the MIMIC-IV dataset is available on \textit{PhysioNet}
\cite{goldberger2000physiobank}, a repository for freely accessible medical data
and tools for computational medicine research. After downloading the collection
of clinical notes, we utilized LLMs to translate the English discharge summaries
to German. Specifically, we employed the multilingual \textit{LLaMA 3.1 8B}
\cite{dubey2024llama3} model in an API-like manner by providing the prompt as
shown in Tab.~\ref{tab:llama}. The translated notes were then included in the
pre-training corpus, consisting of 330,485 documents and 5,310 MB of data.

\begin{table}[htb]
    \centering
    \begin{tabular}{p{0.9\textwidth}}
        \toprule
        \textbf{System Prompt:} \\
        You are an API-like assistant, and output only the plain response
        without further explain or comment the output. \\
        \midrule
        \textbf{User Instruction:} \\
        Translate the following text strictly into German. Do not replace the
        \_\_\_ pseudonymization masks. \texttt{<English Text>} \\
        \bottomrule
    \end{tabular}
    \caption{LLaMA 3.1 system prompt and user instruction used for MIMIC-IV
    translation}
    \label{tab:llama}
\end{table}

\subsubsection{Web Crawl}
To enrich our corpus with current medical content from the German web, a web
crawl was performed using the implementation described in
\cite{deng2025crawler}, which extends the open-source crawler \textsc{Apache
Nutch}~\cite{khare2004nutch}. The crawl was seeded with a combination of domains from the \textit{tala-med search}~\cite{specht2025evaluating} index as well as the seed sources provided by the \textit{sampled German Health Web} (sGHW)~\cite{zowalla2020crawling} project. Tala-med search is a specialized search engine that provides high-quality, evidence-based health information. In its current version, it indexes 26 trustworthy German health websites and ensures strict user privacy. The sGHW project represents previous efforts to index health-related
web content in the German language and employed a specialized focused crawler to
create an index of 22,405 German health websites. The sGHW index was limited to
websites with \texttt{.de}, \texttt{.at}, and \texttt{.ch} top-level domains,
and used a support vector machine to filter content for health relevance
automatically. Our crawl was configured with parameters \texttt{depth=3} and
\texttt{topN=100}. In web crawling, \texttt{depth} refers to the number of hops
or iterations the crawler will follow links from the seed URL, while the width,
called \texttt{topN}, specifies the maximum number of URLs to fetch in each
iteration. These parameters control the crawling process and were chosen to
allow for systematic exploration of linked content while maintaining a
manageable scope. 

Despite the focused seed list, unsuitable as well as nonmedical content, such as
advertisements, remained present in the crawl data due to the nature of web
crawling. To address this issue, we developed a text classifier in order to
filter medical and scientific content from general web content, ensuring the
relevance of the gathered data. The classifier was built by fine-tuning
GeistBERT on a binary-labeled dataset derived from the scientific portion of the
\textit{10kGNAD}~\cite{10kGNAD} corpus. The 10kGNAD dataset is a subset of the
\textit{One Million Posts Corpus} \cite{schabus2017one} and consists of 10,000
German news articles, including 573 focused on scientific topics. These
scientific articles make up the first half of the fine-tuning dataset, while the
second half was created using a stratified sample to ensure a balanced dataset
and that each category was proportionally represented. The classifier's
performance was evaluated using a manually labeled subset of the web crawl data
of size 119, which was annotated using \textsc{Label Studio}~\cite{labelstudio},
an open-source data labeling tool. On this test set, the classifier achieved an
\ff{} score of 80.34\%, indicating a reliable level of accuracy. Following this
validation, we applied the classifier to filter the complete web crawl dataset.
After filtering, we removed documents from the web crawl with less than 40
characters, those containing the Unicode replacement character \texttt{U+FFFD}
due to encoding issues, and duplicates. For the remaining documents, we removed
phone numbers, email addresses, URLs and emojis utilizing the
\textsc{clean-text} \cite{clean-text} Python library. The preprocessing of the
web crawl resulted in a final collection of 93,642 documents and 512 MB of data. Both the classifier~\cite{christbertscignad_tcls_2024} and the scientific subset used for training, referred to as \textit{sciGNAD}~\cite{christbertscignad_2024}, are publicly released to support reproducibility and downstream research.

\subsection{Pre-Training} \label{sec:pre-training}

Leveraging the created large-scale biomedical corpus as described in
Sec.~\ref{sec:corpus} as well as the architectural foundation laid by the
current state-of-the-art German general-purpose language model GeistBERT, we
developed biomedical adaptations by following three main strategies:

\begin{enumerate}
    \item \textbf{Continued pre-training}: Starting from the checkpoint of GeistBERT, we initialize a RoBERTa base model with the identical weights and general domain vocabulary. Subsequently, all parameters of the model are retrained on our 13.5 GB training data as listed in Tab.~\ref{tab:corpus_stats}. Essentially, this approach is equivalent to extending GeistBERT's pre-training dataset with the new data, which is why this strategy is known as \textit{continued} pre-training. The created model following this approach will be referred to as \ChristBERT.\\
    
    \item \textbf{Pre-training from scratch}: We also explored the possibility of pre-training a RoBERTa model from scratch using the same architecture and vocabulary as GeistBERT, but without any initialization from the general domain model. As a result, this model solely learns language representations from our biomedical corpus. We denote this model as \ChristBERT\textsubscript{scratch}.\\

    \item \textbf{Vocabulary adaptation}: In order to study the impact of a domain-specific vocabulary, this strategy involves the creation of a new vocabulary based on the created biomedical corpus and follows the same pre-training process as \ChristBERT\textsubscript{scratch}.
    The vocabulary is generated analogously to GeistBERT, using a GPT-2 style byte pair encoding (BPE) tokenizer with a target vocabulary size of 52,000 tokens. The resulting model is referred to as \ChristBERT\textsubscript{BPE}.
\end{enumerate}

Each of the three models was pre-trained using the \textsc{fairseq}
\cite{ott2019fairseq} framework on the domain-specific corpus presented in
Sec.~\ref{sec:corpus}, which amounts to 13.5 GB of uncompressed text data. The documents comprising the training data, as listed in Tab.~\ref{tab:corpus_stats} were shuffled in order to improve pre-training robustness. The models underwent training for 100,000 update steps with a batch size of 8,192, utilizing weight initialization based on one of the three previously outlined strategies. We adapted GeistBERT's pre-training configuration, which closely aligns with RoBERTa's standard training setup~\cite{liu2019roberta}, encompassing dynamic masking for the WWM learning objective, \textit{AdamW} optimizer parameters, and a fixed sequence length of 512 tokens. To comply with the maximum input sequence length of the model, full sentences from multiple documents in the pre-training corpus were packed into text segments. This procedure allows for retention of natural sentence structure despite the use of fixed-length sequences. For efficient data access, the \textsc{fairseq} library converts the input data into a binary format and utilizes memory-mapped file I/O. A warmup phase of 10,000 iterations was implemented, gradually increasing the learning rate to a maximum of \num{7e-4} for \ChristBERT{} and \num{6e-4} for \ChristBERT\textsubscript{scratch} and \ChristBERT\textsubscript{BPE}, followed by a polynomial decay to zero. The complete pre-training procedure was performed on clusters equipped with either four Nvidia A100 interconnected via SXM or two Nvidia H100 GPUs. The cumulative training time for the three models amounted to approximately 21.7 days (refer to Tab.~S2 in the Supplementary Material).

\subsection{Language Modeling Evaluation} \label{sec:lm_eval}

To assess the impact of different pre-training strategies, we evaluate the intrinsic language modeling performance of our models using \textbf{perplexity}~\cite{bengio2003neural}. Perplexity is a widely used metric that quantifies how well a language model predicts a sequence of words; lower values indicate better generalization and more confident predictions of unseen text.
The perplexity (commonly abbreviated as ppl) of a model $\theta$ on a test set $\mathcal{W}$ is defined as the inverse probability that $\theta$ assigns to $\mathcal{W}$, normalized by the test set length. More formally, for a sequence of $n$ words $w_{1:n} = (w_1, \ldots, w_n)$, the perplexity is given by:

\begin{align} \label{eq:ppl}
    \text{ppl}_{\theta}(w_{1:n}) &= \Pr{}_{\theta}(w_{1:n})^{-\frac{1}{n}} \\
                                 &= \sqrt[n]{\frac{1}{\Pr{}_{\theta}(w_{1:n})}}
\end{align}

We can use the chain rule of probability to express the perplexity of a sequence
of words as the product of the probabilities of each word given its preceding
words:

\begin{equation}
    \text{ppl}_{\theta}(w_{1:n}) = 
    \sqrt[n]{\prod_{i=1}^{n} \frac{1}{\Pr{}_{\theta}(w_i | w_{1:i-1})}}
\end{equation}

Note that due to the inverse relationship in Eq.~\ref{eq:ppl}, higher
probabilities assigned to word sequences correspond to lower perplexity values.
Consequently, a model with lower perplexity indicates that it is a better
predictor of the given test set. Minimizing perplexity is equivalent to
maximizing the probability of the test set as predicted by the LM.

\subsection{Downstream Task Evaluation} \label{sec:fine-tuning}

With minimal adjustments to its architecture, pre-trained LMs can be adapted for downstream applications by fine-tuning them on task-specific
datasets.
Fine-tuning involves adding task-specific layers or adaptation heads that process the model’s hidden representations.
The fine-tuning process consists of continued training using labeled data from supervised datasets to adjust the weights of both the pre-trained model and the task-specific layers added on top.

To demonstrate the efficacy of our domain-adapted model in biomedical language modeling, we fine-tune and evaluate the \ChristBERT{} models on two common biomedical downstream tasks: Named entity recognition (NER) and text classification.

\subsubsection{Named Entity Recognition}
NER is used to extract relevant text spans, such as mentions of diagnoses or medications, from clinical text. We evaluated NER performance on three German biomedical corpora covering oncology and cardiology domains. \textbf{BRONCO150}~\cite{kittner2021bronco150} consists of anonymized sentences from 150 discharge summaries labeled with the categories \textit{Medication}, \textit{Treatment}, and \textit{Diagnosis}. \textbf{GGPONC}~\cite{borchert2022ggponc} is a large-scale corpus derived from German oncology guidelines, containing over 200,000 named entities. There are two major versions of the corpus, from which we used the second version. For our experiments, we selected the most challenging configuration with fine-grained labels and long entity spans to ensure comparability with prior work~\cite{bressem2024medbert}. Finally, \textbf{CARDIO:DE}~\cite{cardiode} comprises 500 cardiovascular discharge letters annotated with six medication-related entity types. We excluded experimental sublabels from CARDIO:DE due to their low inter-annotator agreement.

\subsubsection{Text Classification}

Text classification refers to assigning one or more labels to a document based on its content. We evaluated model performance on two multi-label German biomedical classification tasks:

\textbf{CLEF eHealth 2019}~\cite{crestani2019experimental, clef2019nts, clef2019test} consists of 8,385 German non-technical summaries (NTS) of planned animal studies from the AnimalTestInfo database~\cite{animaltestinfo}. Each summary is annotated with zero or more ICD-10 codes. We followed prior work~\cite{lentzen2022critical} in filtering out rare classes (fewer than 25 occurrences), resulting in 5,688 documents and 230 classes. \textbf{JSynCC}~\cite{lohr2018sharing} contains 867 synthetically generated German case reports from 10 medical textbooks, each annotated with one or more medical specialties. To address class imbalance, we again retained only frequently occurring labels, reducing the dataset to 534 documents and 6 classes, including \textit{Trauma Surgery}, \textit{Anesthesiology}, and \textit{Orthopedics}.

Both datasets exhibit significant label imbalance and are treated as multi-label classification problems.

\subsubsection{Evaluation Metrics}

We report standard classification metrics: precision, recall, and \ff{} score. Following common practice in biomedical NER and multi-label classification~\cite{tjong2003introduction, harbecke2022only}, we used micro-averaged \ff{} as our primary metric to account for class imbalance and capture overall model performance. 

\subsubsection{Dataset Preparation}


For all NER benchmarks, we employed the \textsc{BigBIO}~\cite{fries2022bigbio} library, which provides harmonized dataset schemas, standardized IOB2 entity annotations, and consistent data access tooling for biomedical NLP. For text classification tasks, we used the \textsc{Huggingface Datasets} library~\cite{lhoest2021datasets}, which also serves as a foundation for BigBIO. Whenever available, we preserved the official training, validation, and test splits. For datasets without predefined splits, namely BRONCO150, CARDIO:DE, and JSynCC, we applied stratified random partitioning, allocating 80\%, 10\%, and 10\% of the data to training, validation, and testing, respectively. 
Figure~\ref{fig:benchmark_stats} illustrates the label distribution across splits for each benchmark. We exclude the CLEF eHealth 2019 dataset from this figure, as it contains 230 possible classes.

\begin{figure}[htbp]
    \centering
    \begin{subfigure}{0.49\textwidth}
        \centering

\begin{tabular}{l ccc}
    \toprule
    \multirow{2}{*}[-0.5\dimexpr \aboverulesep + \belowrulesep + \cmidrulewidth]{\bfseries Entity} &
    \multicolumn{3}{c}{\bfseries Split} \\
    \cmidrule(lr){2-4}
    & Train & Val & Test \\
    \midrule
    Diagnosis & 3022 & 368 & 396 \\
    Medication & 1099 & 134 & 160 \\
    Treatment & 2036 & 234 & 263 \\
    \bottomrule
\end{tabular}

        \caption{BRONCO150}
    \end{subfigure}
    \hfill
    \begin{subfigure}{0.49\textwidth}
        \centering

\begin{tabular}{l rrr}
    \toprule
    \multirow{2}{*}[-0.5\dimexpr \aboverulesep + \belowrulesep + \cmidrulewidth]{\bfseries Entity} &
    \multicolumn{3}{c}{\bfseries Split} \\
    \cmidrule(lr){2-4}
    & Train & Val & Test \\
    \midrule
    ActiveIng & 4803 & 699 & 598 \\
    Drug & 1372 & 154 & 154 \\
    Duration & 993 & 124 & 116 \\
    Form & 123 & 20 & 13 \\
    Frequency & 4097 & 596 & 457 \\
    Strength & 4004 & 575 & 447 \\
    \bottomrule
\end{tabular}

        \caption{CARDIO:DE}
    \end{subfigure} 
    \\[\baselineskip]
    \begin{subfigure}{0.8\textwidth}
        \centering

\begin{tabular}{l rrr}
    \toprule
    \multirow{2}{*}[-0.5\dimexpr \aboverulesep + \belowrulesep + \cmidrulewidth]{\bfseries Entity} &
    \multicolumn{3}{c}{\bfseries Split} \\
    \cmidrule(lr){2-4}
    & Train & Val & Test \\
    \midrule
    Clinical Drug & 10973 & 2522 & 2219 \\
    Diagnosis / Pathology & 43249 & 9344 & 9997 \\
    Diagnostic & 17222 & 4035 & 3530 \\
    External Substance & 537 & 181 & 149 \\
    Nutrient / Body Substance & 2462 & 600 & 558 \\
    Other Finding & 31219 & 6520 & 7058 \\
    Therapeutic & 36288 & 7254 & 8421 \\
    \bottomrule
\end{tabular}

        \caption{GGPONC with \textit{fine}-grained entity classes and
        \textit{long} annotation spans}
    \end{subfigure}
    \\[\baselineskip] 
    \begin{subfigure}{0.8\textwidth}
        \centering

\begin{tabular}{l rrr}
    \toprule
    \multirow{2}{*}[-0.5\dimexpr \aboverulesep + \belowrulesep + \cmidrulewidth]{\bfseries Class} &
    \multicolumn{3}{c}{\bfseries Split} \\
    \cmidrule(lr){2-4}
    & Train & Val & Test \\
    \midrule
    Trauma Surgery & 207 & 27 & 32 \\
    Ophthalmology & 31 & 3 & 2 \\
    Orthopedics & 217 & 32 & 33 \\
    Emergency Medicine & 87 & 9 & 11 \\
    Traumatology & 17 & 1 & 1 \\
    Anesthesiology & 30 & 3 & 2 \\
    \bottomrule
\end{tabular}

        \caption{JSynCC}
    \end{subfigure}
    \caption{Entity and class distributions of the downstream tasks}
    \label{fig:benchmark_stats}
\end{figure}

\subsubsection{Experimental Setup}

To evaluate downstream performance, we conducted fine-tuning experiments with hyperparameter optimization on each task. Specifically, we performed a grid search over batch size and learning rate, as detailed in Table~\ref{tab:hyperparams}, yielding 28 trials per task. The search space is based on the GeistBERT evaluation setup~\cite{scheibleschmitt2025geistbertbreathinglifegerman} and extended with additional learning rate values. Each trial used a warmup step ratio of 10\% and trained for up to 30 epochs. The best model checkpoint was selected based on validation set performance. For both NER and classification tasks, we report \textit{micro-averaged} precision, recall, and \ff{} scores on each benchmark's test set.


\begin{table}[htb]
    \centering
    \begin{tabular}{lc}
    \toprule
    \bfseries Parameter & \bfseries Values\\
    \midrule
    Learning Rate & \num{7e-5}, \num{5e-5}, \num{2e-5}, \num{1e-5}, \num{7e-6}, \num{5e-6}, \num{1e-6} \\
    Batch Size & 16, 32, 48, 64 \\
    \bottomrule
\end{tabular}

    \caption{Hyperparameters used in the grid search for the downstream tasks}
    \label{tab:hyperparams}
\end{table}

Unlike perplexity (see Sec.~\ref{sec:lm_eval}), which evaluates intrinsic language modeling ability, downstream task performance enables direct comparison of model efficacy across architectures and domains. To benchmark the \ChristBERT{} models, we selected four state-of-the-art (SOTA) German Transformer-based language models—two domain-specific and two general-purpose baselines (Table~\ref{tab:models}). All model-task combinations underwent identical hyperparameter tuning and evaluation procedures for fair comparison. A brief overview of each baseline is provided below; additional architectural details are listed in Table~S1 in the Supplementary Material.


\begin{table}[htb]
    \centering
    \begin{tabular}{l rrr}
    \toprule
    \bfseries \multirow{2}{*}{Model} & 
    \bfseries \multirow{2}{*}{Type} & 
    \bfseries \multirow{2}{*}{Domain} &
    \bfseries Corpus \\
    & & & \bfseries Size (GB) \\
     \midrule
    \ChristBERT & RoBERTa & Biomedical & 1,482 + 13.5 \\
    \ChristBERT\textsubscript{scratch} & RoBERTa & Biomedical & 13.5 \\
    \ChristBERT\textsubscript{BPE} & RoBERTa & Biomedical & 13.5 \\
    medBERT.de~\cite{bressem2024medbert} & BERT & Biomedical & 10.3 \\
    BioGottBERT~\cite{lentzen2022critical} & RoBERTa & Biomedical & 145 + 0.8 \\
    GeistBERT~\cite{scheibleschmitt2025geistbertbreathinglifegerman} & RoBERTa & General & 145 + 1,337 \\
    GeBERTa~\cite{dada2023impact} & DeBERTa & General & 167 \\
    \bottomrule
\end{tabular}

    \caption[Architecture, domain, and corpus size of evaluated models]{
    Architecture, domain, and corpus size of evaluated models. For \ChristBERT,
    BioGottBERT and GeistBERT, corpus size indicates the size of the initial +
    continuous pre-training corpus.}
    \label{tab:models}
\end{table}

\paragraph{medBERT.de} is based on the BERT~\cite{devlin2019bert} base architecture and is specialized for the German medical domain. Similar to \ChristBERT\textsubscript{BPE}, it was trained from scratch with a custom domain-specific vocabulary on a large and diverse 10.3 GB corpus, comprising 4.7 million German medical documents from eleven different sources, including articles from the German health web, scientific texts, medical books, and real-world clinical data such as electronic health records and radiology reports from Charité University Hospital. This substantial dataset translated into SOTA performance on various medical benchmarks, particularly for longer and more complex texts, such as NER and ICD-10 chapter classification from radiology discharge summaries and surgical reports.


\paragraph{BioGottBERT} is a domain-adapted variant of the unfiltered base version of GottBERT~\cite{scheible2020gottbert}, a RoBERTa-based model trained on the German portion of the OSCAR corpus~\cite{suarez2019asynchronous} (145\,GB of general text).
Similar to \ChristBERT, BioGottBERT was \textit{continuously pre-trained} on 809\,MB of biomedical German texts, specifically from Wikipedia, scientific abstracts and drug leaflets. Despite the small biomedical corpus, BioGottBERT demonstrated notable improvements over its general-domain counterpart on a variety of medical NLP tasks, including NER and classification problems. Our benchmark selection closely follows that of BioGottBERT. The shared tokenizer enables direct comparability between \ChristBERT{} and BioGottBERT across all tasks.


\paragraph{GeistBERT} is a general-domain German language model based on the RoBERTa base architecture. It follows a \textit{continued pre-training} approach, initializing from the best checkpoint of filtered GottBERT~\cite{scheible2020gottbert} (94,530 steps), and extending it with 100,000 further training steps using WWM. The training corpus spans 1.3\,TB of partially deduplicated German data, including crawled web text and publicly accessible legal documents~\cite{nguyen2023culturax, tiedemann2012parallel}. GeistBERT achieves SOTA results across NER, classification, and natural language inference tasks, outperforming even larger models. GeistBERT serves as the general-domain reference for assessing the impact of domain-specific pre-training.


\paragraph{GeBERTa} is another general-domain model that employs the DeBERTa~\cite{he2021deberta} base architecture, featuring \textit{disentangled attention} for improved contextual representation. It was trained on 167\,GB of heterogeneous German data, including formal, informal, legal, medical, and literary text. GeBERTa has been evaluated on general and medical NER, sentiment analysis, hate speech detection, and question answering tasks. As a non-RoBERTa baseline, it allows comparison of architectural effects and cross-domain training data on downstream performance.

\subsubsection{Implementation Details}

All fine-tuning experiments were conducted using the \textsc{Huggingface Transformers}~\cite{wolf2020transformers} Python library and the \textsc{Neural Network Intelligence}~\cite{nni} framework for hyperparameter tuning. The choice of libraries was reinforced by their native support for the dataset implementations~\cite{fries2022bigbio,lhoest2021datasets}. For classification, model inputs were tokenized with truncation to, and padding up to the maximum length of 512 tokens. In the case of NER, longer sequences were split into one or more sequences of 64 tokens except for GGPONC, which was split into 128 tokens. These values correspond to the 95-th percentile of the sequence lengths across each dataset's training, validation and test splits. The evaluation metrics were computed using the \textsc{seqeval}~\cite{seqeval} Python library for NER and the \textsc{sklearn}~\cite{pedregosa2011scikit} Python library for classification. In \textsc{seqeval}, \textit{strict} evaluation mode was applied, measuring both the correctness of the entity boundary and the entity class. All experiments were conducted on consumer-grade hardware, specifically an NVIDIA RTX 3090 GPU with 24 GB VRAM, ensuring reasonable training times for practitioners. The total computation time for all experiments encompassing all 35 model-dataset combinations, amounted to approximately 6.74 days (refer to Tab. S3 in the Supplementary Material).

\section{Results}\label{sec2}

\subsection{Pre-Training Performance}

\begin{figure}[htbp]
  \centering
  \includegraphics[width=0.95\textwidth]{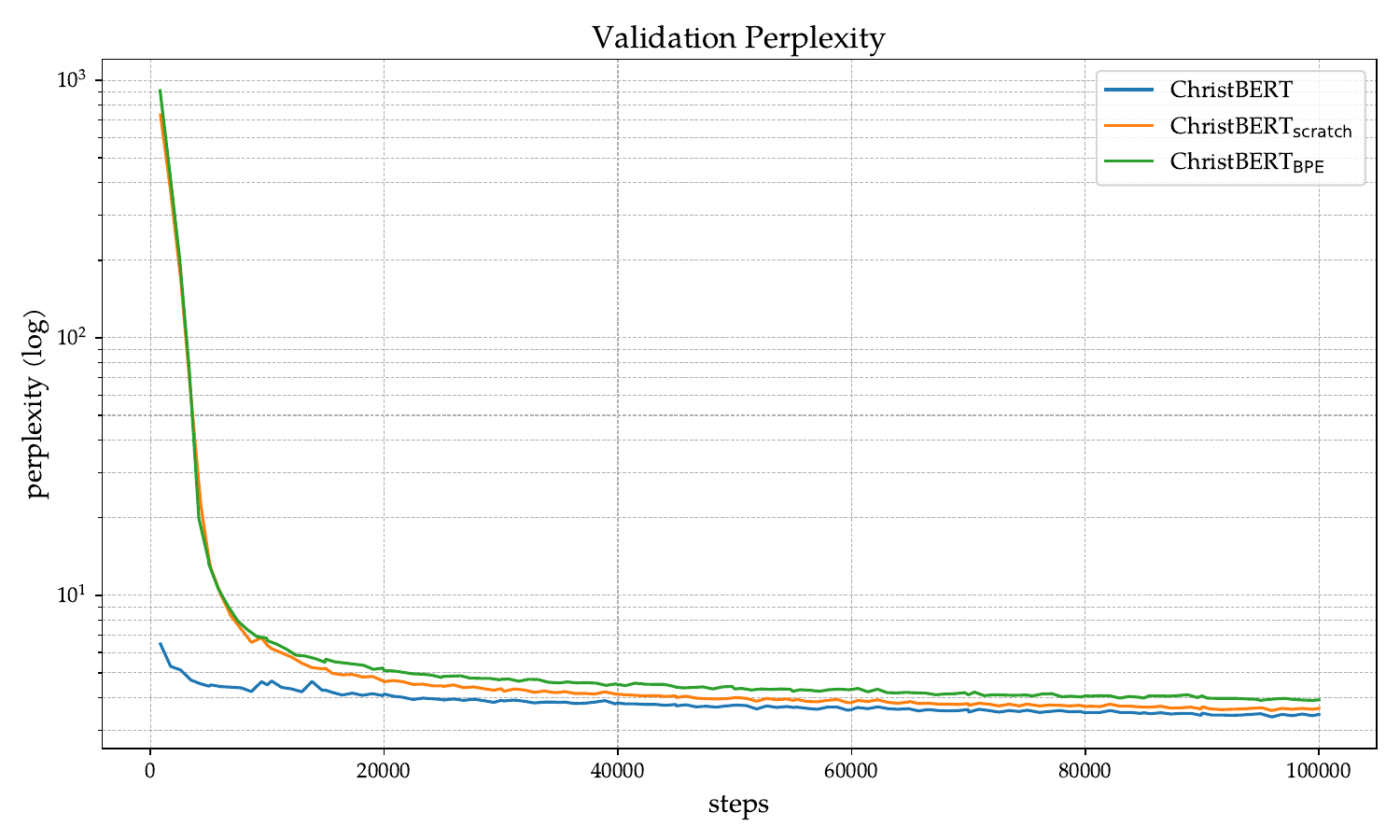}
  \caption[Perplexity during pre-training of \ChristBERT{} models]{Perplexity during pre-training of \ChristBERT{} models. Perplexity is shown in log scale for every optimization step and evaluated on the validation split of the pre-training corpus.}
  \label{fig:ppl}
\end{figure}

Fig.~\ref{fig:ppl} plots the perplexity of the three \ChristBERT{} models during pre-training over 100,000 training steps. Perplexity was evaluated on a held-out validation set of 3,000 randomly chosen documents from our pre-training corpus listed in Tab.~\ref{tab:corpus_stats}. We observe that the different domain adaptation strategies are reflected in the perplexity trajectories in terms of initial perplexity, rate of decline and convergence behavior. 

\subsubsection{Initial Perplexity and Rate of Decline}
Initially, the two \ChristBERT{} variants pre-trained from scratch exhibit high
perplexity values of $57434.5$ and $56343.3$, while the continuously pre-trained
\ChristBERT{} starts with a lower perplexity of $12.64$. The lower perplexity
directly results from the model being initialized with the weights of GeistBERT,
demonstrating the effectiveness of transfer learning. As pre-training
progresses, perplexity decreases steeply during the first 10,000 steps for
\ChristBERT\textsubscript{scratch} and \ChristBERT\textsubscript{BPE}, with the
rate of perplexity reduction following a non-linear pattern across all variants.
The steepest reduction occurs within the first 5,000 steps, where we observe
perplexity values dropping approximately two orders of magnitude from ${\sim}
10^3$ to ${\sim} 10^1$. The observed perplexity curve can be attributed to the
learning rate schedule, in which the learning rate is linearly increased for
10,000 iterations to its maximum. After this warmup phase, the perplexity
trajectory flattens considerably between steps 10,000 and 40,000. Model
perplexity continues to decrease but at a substantially slower rate, which is
due to the learning rate following a polynomial decay to zero after the first
10,000 steps.

\subsubsection{Convergence Behavior}
The continuously pre-trained \ChristBERT{} model converges the fastest,
stabilizing at around a perplexity of 3-4 by 10,000 steps and maintaining the
lowest perplexity throughout pre-training. Additionally, \ChristBERT{}
consistently achieved lower perplexity than both BPE and Scratch variants, with
perplexity values 30-50\% lower during the middle stages of pre-training.
Despite this, after around 40,000-50,000 iterations, all models reach a
relatively stable perplexity level between 2-4. Diminishing returns are observed
after 60,000 steps, suggesting extended pre-training offers minimal improvements
and convergence is achieved. Moreover, we observed divergence in some
pre-training runs, particularly due to high learning rates where model
parameters were updated too aggressively. This divergence manifested as sudden
spikes in perplexity and subsequent failure to converge is shown in
Fig.~S1 in the Supplementary Material. To mitigate divergence, we found that lowering
the learning rate was effective in stabilizing pre-training. This was necessary
for \ChristBERT\textsubscript{scratch} and \ChristBERT\textsubscript{BPE}, where
we reduced the maximum learning rate from \num{7e-4} to \num{6e-4}, while with
the GeistBERT initialization it was possible to use a higher peak learning rate.

\ChristBERT\textsubscript{BPE} consistently demonstrates the highest perplexity values among the three variants, particularly during the middle phase of pre-training. This effect likely stems from its custom byte-pair encoding vocabulary, where it spends the middle stages learning different language representations. As discussed in Sec.~\ref{sec:lm_eval}, perplexity comparisons are most meaningful between models sharing the same tokenizer; therefore this difference does not necessarily indicate inferior model quality.
Moreover, an improvement in an intrinsic measure such as perplexity does not necessarily correlate with enhanced performance in extrinsic measures such as practical downstream language tasks. Nevertheless, perplexity remains a useful proxy for estimating a model’s generalization capacity and its potential effectiveness on downstream tasks.

\subsection{Fine-Tuning Performance}

\subsubsection{Named Entity Recognition}

Tab.~\ref{tab:eval_ner_metrics} shows the performance results of medical
named entity recognition on the BRONCO150, CARDIO:DE and GGPONC datasets.
Detailed results for each entity type in the respective dataset are reported in
Tab.~S7-S9 in the Supplementary Material. The
\ChristBERT{} models consistently outperform the baseline models across all
datasets, establishing a new state-of-the-art German biomedical NER.

\begin{table}[htbp]
    \centering

\begin{tabular}{l ccc ccc ccc}
    \toprule
    \multirow{2}{*}[-0.5\dimexpr \aboverulesep + \belowrulesep + \cmidrulewidth]{\bfseries Model} &
    \multicolumn{3}{c}{\bfseries BRONCO150} &
    \multicolumn{3}{c}{\bfseries CARDIO:DE} &
    \multicolumn{3}{c}{\bfseries GGPONC} \\
    \cmidrule(lr){2-4} \cmidrule(lr){5-7} \cmidrule(lr){8-10}
    & Prec. & Rec. & \ff &
    Prec. & Rec. & \ff &
    Prec. & Rec. & \ff \\
    \midrule
     \ChristBERT & 81.42 & 81.77 & 81.87 & 85.58 & 89.65 & 87.57 & 75.65 & \textbf{79.83} & \textbf{77.69} \\
     \ChristBERT\textsubscript{scratch} & \underline{81.87} & \underline{82.32} & \underline{82.09} & 88.38 & 89.89 & 89.13 & \underline{76.54} & \underline{77.56} & \underline{77.05} \\
     \ChristBERT\textsubscript{BPE} & \textbf{85.71} & \textbf{83.78} & \textbf{84.74} & \underline{89.50} & \textbf{91.31} & \textbf{90.40} & \textbf{76.59} & 77.42 & 77.00 \\
     medBERT.de & 78.67 & 79.58 & 79.12 & 87.66 & 90.02 & 88.83 & 73.89 & 75.78 & 74.73 \\
     BioGottBERT & 76.96 & 78.45 & 77.70 & 88.37 & \underline{90.74} & 89.54 & 75.24 & 75.40 & 75.32 \\ 
     GeistBERT & 75.65 & 79.83 & 77.69 & 85.58 & 89.65 & 87.57 & 74.57 & 75.36 & 74.96 \\
     GeBERTa & 78.67 & 79.58 & 79.12 & \textbf{90.51} & 90.23 & \underline{90.37} & 75.96 & 76.93 & 76.45 \\
    \bottomrule
\end{tabular}
 
    \caption[Overview of micro averaged precision, recall and \ff{} scores
    achieved on the NER tasks]{Overview of micro averaged precision (Prec.),
    recall (Rec.) and \ff{} scores on the NER tasks. All results are shown in
    percent and assess each model's best fine-tuned performance on each
    downstream task's test set. The best model was selected out of 28 runs based
    on its validation set performance. Best score in bold and second best
    underlined.}
    \label{tab:eval_ner_metrics}
\end{table}

On the BRONCO150 dataset, \ChristBERT\textsubscript{BPE} achieves the highest
precision (85.71\%), recall (82.32\%) and \ff{} score (84.74\%), forming a
substantial improvement over both specialized medical models and general
language models. \ChristBERT\textsubscript{scratch} places second with an \ff{}
of 83.33\%, followed closely by \ChristBERT{} with 81.87\%. The performance
delta between \ChristBERT{} variants and other models is particularly evident
when comparing against the general language models. For instance, the \ff{}
score of \ChristBERT\textsubscript{BPE} with 84.74\% represents a 5.62
percentage point improvement over GeistBERT (77.69\%) and a 5.08 percentage
point improvement over GeBERTa (79.12\%). This significant performance gap
underscores the value of domain-specific pre-training for NER in medical texts.

The performance on the CARDIO:DE dataset presents a different pattern of
results. Here, all compared models showed more similar performances, with
\ChristBERT\textsubscript{BPE} performing the best, followed closely by GeBERTa,
which leads among the baselines and demonstrates on par NER efficacy. Both
mentioned models achieve high \ff{} scores of 90.40\% and 90.37\%, respectively,
differing in precision and recall. This dataset highlights the potential for
general language models to perform competitively in certain medical subdomains
when trained appropriately.

The GGPONC dataset presents the most challenging evaluation scenario with eight
fine-grained semantic classes and long entity spans across a large corpus of
oncology documentation. On this complex dataset, \ChristBERT{} models again
demonstrate superior performance compared to the baseline models, with
\ChristBERT{} achieving the highest recall (79.83\%), while
\ChristBERT\textsubscript{BPE} attains the highest precision (76.59\%). Here,
\ChristBERT\textsubscript{BPE} and \ChristBERT\textsubscript{scratch} match each
other's precision, recall and \ff{} scores. The performance advantage of our
pre-training corpus on GGPONC is particularly noteworthy given the complexity of
this dataset. With an \ff{} of 77.69\%, \ChristBERT{} is the best performing NER
model and outperforms the next best non-\ChristBERT{} model GeBERTa at 76.45\%
by 1.24 percentage points. The demonstrated advantage in the most complex
dataset suggests that the domain-specific pre-training of \ChristBERT{} models
enables more effective learning of the nuanced entity boundaries and semantic
distinctions required for fine-grained medical entity recognition.

\subsubsection{Text Classification}

Tab.~\ref{tab:eval_cls_metrics} presents the classification results for each
model on the CLEF and JSynCC classification datasets. Detailed results for each
topic category in JSynCC can be found in Tab.~S9 in the Supplementary Material.
We omit a separate per class drill-down for the CLEF dataset as it contains over
230 classes. As such, the CLEF benchmark poses the more challenging multi-label
classification task, while JSynCC only requires assigning labels out of six
medical categories.

\begin{table}[htbp]
    \centering

\begin{tabular}{l ccc ccc}
    \toprule
    \multirow{2}{*}[-0.5\dimexpr \aboverulesep + \belowrulesep + \cmidrulewidth]{\bfseries Model} &
    \multicolumn{3}{c}{\bfseries CLEF} &
    \multicolumn{3}{c}{\bfseries JSynCC} \\
    \cmidrule(lr){2-4} \cmidrule(lr){5-7} 
    & Prec. & Rec. & \ff & 
    Prec. & Rec. & \ff \\
    \midrule
    \ChristBERT & 78.12 & 75.34 & 76.03 & 89.01 & \textbf{100} & \underline{94.19} \\
    \ChristBERT\textsubscript{scratch} & \textbf{93.68} & 85.17 & \underline{89.22} & \underline{91.86} & 97.53 & \textbf{94.61} \\
    \ChristBERT\textsubscript{BPE} & 88.22 & \underline{88.35} & 88.28 & 89.53 & 95.06 & 92.22 \\
    medBERT.de & 89.21 & 87.59 & 88.40 & 91.25 & 90.12 & 90.68 \\
    BioGottBERT & 88.30 & 87.90 & 88.10 & 88.89 & \underline{98.77} & 93.57 \\
    GeistBERT & \underline{90.43} & 72.92 & 80.74 & \textbf{92.59} & 92.59 & 92.59 \\
    GeBERTa & 88.91 & \textbf{89.71} & \textbf{89.31} & \textbf{92.59} & 92.59 & 92.59 \\
    \bottomrule
\end{tabular}
 
    \caption[Overview of micro averaged precision, recall and \ff{} scores
    achieved on the classification tasks]{Overview of micro averaged precision
    (Prec.), recall (Rec.) and \ff{} scores on the classification tasks. All
    results are shown in percent and assess each model's best fine-tuned
    performance on each downstream task's test set. The best model was selected
    out of 28 runs based on its validation set performance. Best score in bold
    and second best underlined.}
    \label{tab:eval_cls_metrics}
\end{table}

On the CLEF dataset, GeBERTa achieves the highest \ff{} score at 89.31\%, driven
by its superior recall at 89.71\%. Nonetheless,
\ChristBERT\textsubscript{scratch} demonstrates the highest precision (93.68\%),
indicating that it is more effective at minimizing false positives. However, its
recall (85.17\%) is lower than GeBERTa's, resulting in a lower overall \ff{} at
89.22\%. To our surprise, we observe that both general and domain-specific
models perform similarly on this dataset. Notably, the continuously pre-trained
\ChristBERT{} variant shows the lowest overall performance among the evaluated
models on this dataset with an \ff{} of 76.03\%. Its performance differs by 4.61
percentage points from the next best model GeistBERT (80.64\%), its general
domain counterpart. This suggests that the continuous pre-training approach may
not be as effective for complex multi-label classification problems,
particularly when compared to the other \ChristBERT{} variants, which were
pre-trained with the same corpus but with different initialization strategies. 

It should be noted that GeBERTa included CLEF data in its pre-training corpus,
meaning it had already seen this data before evaluation. This might
explain its exceptionally high performance compared to other models and should
be considered when interpreting these results. Even so, medBERT.de achieves
strong performance with an \ff{} score of 88.40\%, demonstrating that domain
adaptation across different medical subdomains supports the processing of
specialized terminology and concepts in animal experiment documentation.

On the JSynCC dataset, the majority of \ChristBERT{} models considerably
outperform the baseline models, with \ChristBERT\textsubscript{scratch}
achieving the highest \ff{} score of 94.61\%, closely followed by \ChristBERT{}
at 94.19\% and a shared third place between GeistBERT and GeBERTa at 92.59\%. A
particularly striking observation is the perfect recall (100\%) of \ChristBERT{}
on the JSynCC dataset, indicating that it identifies all relevant specialty
classifications across the test documents. However, its precision (89.01\%) is
lower than other models, resulting in an \ff{} of 94.19\%. This pattern suggests
that \ChristBERT{} may be over-predicting certain class labels, but its
comprehensive coverage ensures no relevant classifications are missed, a
characteristic that could be valuable in clinical applications where missing a
relevant specialty category might have significant consequences. 

The performance clustering on JSynCC is notably tight, with all models achieving
\ff{} scores between 92.59\% and 94.61\%. Notably, BioGottBERT achieves the
second-highest overall performance on JSynCC with an \ff{} of 93.57\% and recall
of 98.77\%. This suggests that the synthetic nature of this corpus may present
more standardized linguistic patterns that various model architectures can
effectively learn during fine-tuning. Furthermore, while
\ChristBERT\textsubscript{BPE} has consistently shown the best performance in
NER tasks, it does not rank among the top models on all classification
benchmarks. This indicates that the BPE vocabulary may not be as effective for
text classification tasks, where the model's ability to generalize across
different contexts and semantic meanings is crucial.

\subsubsection{Cross-Model Analysis and Domain Specialization Effects}
Among the \ChristBERT{} variants, \ChristBERT\textsubscript{BPE} consistently
demonstrates strong performance across all NER datasets, achieving the highest
or second-highest \ff{} scores in each experiment. This suggests that the custom
BPE vocabulary approach may offer advantages for handling the morphological
complexity and specialized vocabulary found in German medical texts. Despite its
seemingly weaker performance during pre-training as indicated by higher
perplexity values, its downstream performance confirms that pre-training metrics
do not necessarily translate into task-specific effectiveness.

\ChristBERT\textsubscript{scratch} also performs competitively across NER
datasets, indicating that domain-specific training from initialization can be
effective without leveraging transfer learning from general domain pre-training.
The continuously pre-trained \ChristBERT{} model shows particular strength in
the GGPONC dataset, suggesting it may have advantages for handling complex,
fine-grained entity recognition tasks.

The comparison between specialized medical models (\ChristBERT{} variants,
medBERT.de, BioGottBERT) and general language models (GeistBERT, GeBERTa)
reveals distinct performance behavior in medical NER. In BRONCO150 and GGPONC,
domain-specific models generally outperform general models, confirming the value
of specialized pre-training for oncology text. However, in CARDIO:DE, GeBERTa
achieves the highest \ff, suggesting that general language models can be
competitive in certain medical subdomains when trained on heterogeneous and
cross-domain data. Notably, 8\% of GeBERTa’s pre-training data consisted of
medical texts. 

This variability illustrates that domain specificity presents different
advantages depending on the particular medical subdomain and entity types being
targeted. The general language models appear more competitive on CARDIO:DE,
possibly due to differences in writing style, terminology standardization, or
entity class definitions between cardiovascular and oncology domains.
Interestingly, we observe GeistBERT exhibiting equivalent performance to the
domain-adapted model BioGottBERT. We attribute this mainly to the relatively
small size of BioGottBERT's biomedical training corpus (0.8 GB), highlighting
the importance of corpus size in achieving effective domain adaptation.

An analysis of precision and recall values reveals different optimization
patterns across models. \ChristBERT\textsubscript{BPE} tends to favor precision
over recall in BRONCO150 and GGPONC, while achieving high values in both metrics
for CARDIO:DE. In contrast, the continuously pre-trained \ChristBERT{} shows
stronger recall performance, particularly in GGPONC. These trade-offs have
important implications for clinical applications, where the relative importance
of precision versus recall may vary based on the specific use case.

For the classification tasks, a complementary pattern emerges. While \ChristBERT\textsubscript{BPE} dominated in NER, it was outperformed by \ChristBERT\textsubscript{scratch} and the baseline GeBERTa on both CLEF and JSynCC. This suggests that the advantages of byte-pair encoding may not generalize equally across all task types. In contrast, \ChristBERT\textsubscript{scratch} delivered consistently strong results in both precision and recall, particularly excelling on JSynCC, which implies that full pre-training on domain-specific corpora enables robust feature representations for document-level tasks.

The continuously pre-trained \ChristBERT{} variant showed the weakest classification performance, likely due to residual biases from general-domain pre-training interfering with adaptation to complex, multi-label classification setups like CLEF. Interestingly, despite its poor performance on CLEF, this variant achieved perfect recall on JSynCC, underscoring that continued pre-training can support comprehensive label coverage but may lead to over-prediction and reduced precision.

\section{Discussion}
\subsection{General Findings} \label{sec:findings}

In this study, we systematically explored three complementary strategies for
domain adaptation of German biomedical language models: continued pre-training
from a general-domain model (\ChristBERT{}), pre-training from scratch
(\ChristBERT\textsubscript{scratch}), and vocabulary adaptation via
domain-specific subword tokenization (\ChristBERT\textsubscript{BPE}). All
models were pre-trained on a newly curated 13.5 GB biomedical corpus and
evaluated on downstream biomedical tasks, including NER and text classification. Our experiments reveal several principal findings
regarding the three investigated domain adaptation strategies.

First, continued pre-training proved particularly effective in terms of
efficiency. \ChristBERT{} achieved the lowest perplexity and converged fastest,
underscoring the benefits of leveraging general-domain knowledge. This
advantage, however, did not always translate into downstream superiority for
initializing biomedical models. Instead, its performance varied among downstream
tasks: While \ChristBERT{} excelled in NER, particularly on GGPONC, it ranked
lowest on complex classification tasks such as CLEF, indicating that inherited
general-domain priors may not always be beneficial for complex classification
tasks.

Second, pre-training from scratch led to robust and often superior downstream
performance. \ChristBERT\textsubscript{scratch} achieved top results on text
classification tasks, particularly JSynCC, where it attained the highest \ff{}
score. This suggests that domain-exclusive representations learned from scratch
may offer advantages in classification scenarios requiring broader semantic
coverage and contextual generalization.

Third, domain-specific vocabulary adaptation (\ChristBERT\textsubscript{BPE})
yielded the strongest performance for entity-centric tasks. Despite higher
perplexity during pre-training, this variant excelled in NER tasks across all
datasets, achieving state-of-the-art results on BRONCO150 and CARDIO:DE.
However, its performance in classification tasks was less competitive,
indicating that the benefits of domain-optimized tokenization are most
pronounced in tasks sensitive to terminological precision and morphological
complexity.

Finally, comparisons to general-purpose language models highlighted the
importance of domain adaptation. While general models such as GeistBERT and
GeBERTa remained competitive on certain datasets like CARDIO:DE and CLEF, they
were consistently outperformed by the \ChristBERT{} variants on more specialized
or complex biomedical tasks. Furthermore, smaller-scale domain adaptation
efforts (e.g., BioGottBERT) could not match the performance gains achieved
through our larger corpus and comprehensive pre-training strategies.

In summary, our findings emphasize that no single adaptation strategy
universally outperforms the others. Continued pre-training offers rapid
convergence and strong generalization. From-scratch pre-training provides robust
performance for classification, while additional domain-specific vocabulary is
most beneficial for specialized tasks like NER. Our results highlight that the
suitability of domain-specific tokenization strategies, such as a custom BPE
vocabulary, is highly task-dependent. This suggests that domain-specific BPE
tokenization is especially beneficial for entity recognition, where accurate
boundary detection and handling of rare terms are critical. In contrast,
classification tasks often rely more on the model's ability to generalize over
broader semantic and syntactic patterns rather than fine-grained tokenization.
Thus, in such contexts, the rigid subword splits introduced by domain-specific
BPE may offer less benefit, or even introduce unnecessary complexity. These
observations emphasize the importance of aligning vocabulary adaptation
strategies not only with the domain but also with the linguistic properties and
demands of the target task.

\subsection{Findings in the Context of Prior Work} \label{sec:prior_work}

Our findings align well with and extend previous work on domain-adaptive
pre-training. The study~\cite{gururangan2020don} demonstrated that continued
pre-training yields significant gains for domain-specific tasks, especially when
the target domain is distant from the original pre-training corpus. Our results
confirm this for German biomedical NLP: continued pre-training (\ChristBERT) led
to rapid convergence and strong performance in complex NER tasks like GGPONC.
Furthermore, previous findings from~\cite{el2022re} suggest that training from
scratch can be competitive with, or even outperform, continued pre-training on
biomedical classification tasks. Our \ChristBERT\textsubscript{scratch} model
demonstrated this by excelling on both the JSynCC and CLEF classification
benchmarks. In their experiments, the authors of~\cite{el2022re} also observed
that medical-specific vocabularies lead to performance gains in downstream
domain tasks. Our results mirror these prior observations, with
\ChristBERT\textsubscript{BPE} achieving top results in NER, reinforcing the
idea that domain-aligned vocabulary improves handling of specialized
terminology.

Inspired by~\cite{edunov2018understanding}, we translated English medical texts
into German to address the scarcity of native-language biomedical corpora. This
strategy proved effective in terms of downstream task performance compared to
medBERT.de~\cite{bressem2024medbert}, which relied exclusively on original
German data. GeBERTa~\cite{dada2023impact}, which also leveraged translated
medical texts, achieved similarly strong results, particularly in classification
scenarios. Notably, even general-purpose models performed competitively on
classification tasks, underscoring that large-scale general-domain pre-training
enriched with some biomedical content remains a viable approach for such tasks.
Nevertheless, our findings support the approach of translation-based corpus
construction, especially for tasks like biomedical NER, where domain-specific
nuances and terminology require targeted representation learning and original
German resources remain limited.

While implementing the translation strategy for MIMIC-IV with LLaMA 3.1 and
Pubmed Central with NLLB 200, similar to~\cite{dada2023impact} we also observed
that the quality of the machine-translated data was sensitive to translation
settings. In particular with NLLB 200, we noticed that larger context sizes and
sequence lengths frequently resulted in degraded translation quality. Phenomena
such as stuttering and incoherent phrase repetition became evident, especially
in complex biomedical sentences. This degradation can stem from several factors
inherent to current translation models and LLMs used for translation. For
instance, generic LLMs, if configured incorrectly during inference (e.g.
insufficient context window sizes), may fail to attend to the entire input
sequence, effectively \textit{forgetting} earlier parts of the input and
producing incomplete or nonsensical translations. Likewise, many dedicated
translation models are trained on sentences. Consequently, their ability to
handle longer sequences degrades, as the positional embeddings beyond the
trained length are less reliable, leading to instability and errors in
translation~\cite{costa2022no}. To ensure the reliability of the translated
corpus, we therefore opted for a context size of 384 tokens for NLLB 200, which
offered a favorable balance between translation throughput and linguistic
accuracy, mitigating some of these input length-related issues.

\subsection{Limitations and Future Work} \label{sec:limitations}

While this study provides valuable insights into domain adaptation strategies
for German biomedical language models, several limitations remain and point to
promising directions for future research.

Our investigation was limited to the RoBERTa architecture, following the design
path of GeistBERT~\cite{scheibleschmitt2025geistbertbreathinglifegerman} and GottBERT~\cite{scheible2020gottbert}. Although this
ensured comparability, alternative Transformer architectures, including the
recently introduced ModernBERT~\cite{warner2024smarter}, may offer performance
advantages in terms of computational efficiency and input size. RoBERTa's maximum input size
limitation becomes particularly constraining in biomedical contexts, where
clinical documents such as patient records or scientific articles often involve
extended contexts. Future work should explore long-context Transformers.
Architectures such as Longformer~\cite{beltagy2020longformer} and
Nyströmformer~\cite{xiong2021nystromformer} could offer significant advantages
in tasks requiring document-level understanding or the resolution of complex
cross-sentence dependencies \cite{shalumov2023herorobertalongformerhebrew}.

Furthermore, our findings indicate that training from scratch can be most
effective under certain conditions. This discrepancy highlights the need for
further investigation into the factors that influence the effectiveness of
training from scratch versus continued pre-training. Future work should focus on
exploring the specific scenarios and downstream tasks where training from
scratch might outperform continued pre-training. It would be valuable to conduct
a more detailed analysis of the trade-offs between computational resources,
training time, and performance gains. Understanding these dynamics could provide
insights into optimizing model training strategies for various applications.
Similarly, while our models build directly on GeistBERT and GottBERT in terms of
tokenizer design and vocabulary size, these decisions were not revisited for the
biomedical domain. Given the distinct lexical properties of medical language,
alternative vocabulary sizes or tokenization schemes might further optimize
model performance.

Moreover, the range of biomedical benchmarks used, while diverse, does not fully
reflect the variety of clinical language processing needs. Tasks involving
decision support, complex narratives, and clinical reasoning were
underrepresented, which will hopefully change in the near future with the
release of the German Medical Text Corpus
Project~\cite{meineke2023announcement}. Addressing these gaps is important,
especially in light of models like medBERT.de~\cite{bressem2024medbert}, which
explicitly targeted such scenarios. In addition, subdomains such as radiology,
psychiatry, and primary care were not systematically explored, limiting our
conclusions about generalizability.

Our corpus design also presents limitations. While our approach exclusively used
biomedical data, models like GeBERTa~\cite{dada2023impact} have demonstrated
that mixed-domain corpora can enhance generalization, particularly for tasks
that bridge specialized and general language. Investigating mixed corpus
strategies within the same RoBERTa architecture could therefore provide deeper
insights into optimal corpus design for domain-adaptive pre-training. Further
considerations result from performing translation to augment the pre-training
corpus. Although translation enabled the creation of a large biomedical corpus,
the quality of this synthetic data was not manually verified by healthcare
professionals and, as discussed, was sensitive to context size, with larger
sizes impairing coherence. Future work could investigate the effects of
translation quality and translation model behavior itself to assess and mitigate
such artifacts more systematically. Likewise, we did not systematically analyze
the individual contributions of the different data sources within our corpus; it
remains unclear to what extent the translated data specifically improved
performance compared to relying solely on the original German sources.
Additionally, de-identified datasets, i.e. MIMIC-IV, contain artifacts such as
anonymization masks, which are not typically found in natural prose, potentially
affecting performance on other types of text. As a byproduct of the translation
effort, we have obtained a large bilingual corpus of clinical texts (MIMIC-IV
Notes) and biomedical literature (PubMed Central). This corpus could be used to fine-tune German–English translation models for the biomedical domain, supporting both direct clinical applications and future corpus creation. 

Lastly, one should be cautious when interpreting results in cases of potential data leakage. In the case of GeBERTa, the pre-training corpus included CLEF data (without labels), which may still confer an advantage in classification tasks involving this benchmark. In contrast, medBERT.de was pre-trained on GGPONC, which is also part of our NER evaluation. However, medBERT.de performed worse than several models without similar data leakage, suggesting that this exposure did not translate into a measurable advantage in case of NER. This underlines the importance of careful dataset curation and transparency when reporting benchmark results, while also showing that plain-text overlap alone does not guarantee performance gains. 
It remains unclear whether, and to what extent, such data leakage impact downstream performance, especially since both CLEF and GGPONC are relatively small compared to the full pre-training corpora. Determining their exact influence would require dedicated experiments, but we highlight the potential of such effects for further investigation.

\section{Conclusion} \label{chap:conclusion}

This study systematically explored domain adaptation strategies for German biomedical language models: continued pre-training from a general-domain model, training from scratch on biomedical data, and adapting the tokenizer with domain-specific BPE. Central to this effort was the creation of a large-scale pre-training corpus, enriched through translation-based data augmentation to address the scarcity of German clinical text.

Three models were trained using these strategies and benchmarked against existing general and medical German models. Evaluations included intrinsic perplexity and extrinsic performance across five NER and classification datasets. The ChristBERT models achieved state-of-the-art results in 4 of 5 tasks of our setup, though no single strategy consistently outperformed the others. Continued pre-training proved efficient and strong on certain NER tasks; training from scratch excelled in classification; BPE adaptation offered nuanced gains, particularly for specialized terminology. Based on our evaluations, the optimal adaptation strategy depends on task requirements and resource constraints.

This work contributes state-of-the-art German biomedical language models and provides valuable insights into domain adaptation strategies, paving the way for future advancements in clinical text processing and mining. All models including some resources are publicly released to support continued research and application.

\backmatter





\bmhead{Acknowledgements}

The authors gratefully acknowledge the scientific support and resources of the AI service infrastructure LRZ AI Systems provided by the Leibniz Supercomputing Centre (LRZ) of the Bavarian Academy of Sciences and Humanities (BAdW), funded by Bayerisches Staatsministerium für Wissenschaft und Kunst (StMWK).
The authors gratefully acknowledge the resources on the LiCCA HPC cluster of the University of Augsburg, co-funded by the Deutsche Forschungsgemeinschaft (DFG, German Research Foundation) – Project-ID 499211671.
We would like to thank hpsmedia, especially Andreas Lauterbach, for their data contribution, and the authors of medBERT.de, in particular Keno Bressem, for their assistance regarding certain areas of the corpus. We are also grateful to Richard Zowalla for his helpful communication concerning the sGHW project and his openness in sharing insights. Furthermore, we thank Karen Luna Samanez for providing an initial code base for web data deduplication, which supported the corpus preparation for this work.

\bmhead{Data availability}
The pretraining corpus consists of publicly available and licensed biomedical sources, including open-access medical literature, de-identified clinical notes, and curated web data. Redistribution may be restricted for some datasets due to licensing constraints. All resulting models are available on Huggingface.

\bmhead{Materials availability}
The pre-trained \ChristBERT{} models are publicly available at \url{https://huggingface.co/ChristBERT}. Fairseq checkpoints can be provided upon request.

\bmhead{Consent for publication}
Not applicable.

\bmhead{Code availability}
The evaluation framework used in this study is publicly available at:
\url{https://gitlab.com/christbert/evaluation/}.



\bmhead{Conflict of interest}
The authors declare that they have no competing interests.

\bmhead{Author contribution}
Conceptualization, Raphael Scheible-Schmitt;  
Data curation, Henry He, Raphael Scheible-Schmitt and Johann Frei;  
Formal analysis, Henry He;  
Investigation, Henry He and Raphael Scheible-Schmitt;  
Methodology, Henry He and Raphael Scheible-Schmitt;  
Project administration, Raphael Scheible-Schmitt;  
Resources, Johann Frei and Raphael Scheible-Schmitt;  
Software, Henry He, Johann Frei and Raphael Scheible-Schmitt;  
Supervision, Raphael Scheible-Schmitt;  
Validation, Henry He;  
Visualization, Henry He and Raphael Scheible-Schmitt;  
Writing – original draft, Henry He,Johann Frei and Raphael Scheible-Schmitt;  
Writing – review and editing, Henry He, Johann Frei and Raphael Scheible-Schmitt.
All authors read and approved the final manuscript.

\bmhead{Ethics approval and consent to participate}
Not applicable.

\bmhead{Funding}
Not applicable.





\bigskip





\clearpage
\newpage
\bibliography{sn-bibliography}

@inproceedings{scheible2020gottbert,
    title = "{G}ott{BERT}: a pure {G}erman Language Model",
    author = "Scheible, Raphael  and
      Frei, Johann  and
      Thomczyk, Fabian  and
      He, Henry  and
      Tippmann, Patric  and
      Knaus, Jochen  and
      Jaravine, Victor  and
      Kramer, Frank  and
      Boeker, Martin",
    booktitle = "Proceedings of the 2024 Conference on Empirical Methods in Natural Language Processing",
    month = nov,
    year = "2024",
    address = "Miami, Florida, USA",
    publisher = "Association for Computational Linguistics",
    pages = "21237--21250",
}

@article{bressem2024medbert,
  title={Medbert. de: A comprehensive german bert model for the medical domain},
  author={Bressem, Keno K and Papaioannou, Jens-Michalis and Grundmann, Paul and Borchert, Florian and Adams, Lisa C and Liu, Leonhard and Busch, Felix and Xu, Lina and Loyen, Jan P and Niehues, Stefan M and others},
  journal={Expert Systems with Applications},
  volume={237},
  pages={121598},
  year={2024},
  publisher={Elsevier}
}

@article{kittner2021bronco150,
  title={Annotation and initial evaluation of a large annotated German oncological corpus},
  author={Kittner, Madeleine and Lamping, Mario and Rieke, Damian T and G{\"o}tze, Julian and Bajwa, Bariya and Jelas, Ivan and R{\"u}ter, Gina and Hautow, Hanjo and S{\"a}nger, Mario and Habibi, Maryam and others},
  journal={JAMIA open},
  volume={4},
  number={2},
  pages={ooab025},
  year={2021},
  publisher={Oxford University Press}
}

@inproceedings{borchert2022ggponc,
  title={GGPONC 2.0-the German clinical guideline corpus for oncology: Curation workflow, annotation policy, baseline NER taggers},
  author={Borchert, Florian and Lohr, Christina and Modersohn, Luise and Witt, Jonas and Langer, Thomas and Follmann, Markus and Gietzelt, Matthias and Arnrich, Bert and Hahn, Udo and Schapranow, Matthieu-P},
  booktitle={Proceedings of the Thirteenth Language Resources and Evaluation Conference},
  pages={3650--3660},
  year={2022}
}

@article{frei2023gptnermed,
  title={Annotated dataset creation through large language models for non-english medical NLP},
  author={Frei, Johann and Kramer, Frank},
  journal={Journal of Biomedical Informatics},
  volume={145},
  pages={104478},
  year={2023},
  publisher={Elsevier}
}

@article{liu2019roberta,
  title={Roberta: A robustly optimized bert pretraining approach},
  author={Liu, Yinhan},
  journal={arXiv preprint arXiv:1907.11692},
  year={2019}
}

@inproceedings{devlin2019bert,
  title={BERT: Pre-training of Deep Bidirectional Transformers for Language Understanding},
  author={Devlin, Jacob and Chang, Ming-Wei and Lee, Kenton and Toutanova, Kristina},
  booktitle={Proceedings of the 2019 Conference of the North American Chapter of the Association for Computational Linguistics: Human Language Technologies, Volume 1 (Long and Short Papers)},
  pages={4171--4186},
  year={2019}
}

@inproceedings{ott2019fairseq,
  title={fairseq: A Fast, Extensible Toolkit for Sequence Modeling},
  author={Ott, Myle and Edunov, Sergey and Baevski, Alexei and Fan, Angela and Gross, Sam and Ng, Nathan and Grangier, David and Auli, Michael},
  booktitle={Proceedings of the 2019 Conference of the North American Chapter of the Association for Computational Linguistics (Demonstrations)},
  pages={48--53},
  year={2019}
}

@article{dubey2024llama3,
  title={The llama 3 herd of models},
  author={Dubey, Abhimanyu and Jauhri, Abhinav and Pandey, Abhinav and Kadian, Abhishek and Al-Dahle, Ahmad and Letman, Aiesha and Mathur, Akhil and Schelten, Alan and Yang, Amy and Fan, Angela and others},
  journal={arXiv preprint arXiv:2407.21783},
  year={2024}
}

@article{zowalla2020crawling,
  title={Crawling the german health web: Exploratory study and graph analysis},
  author={Zowalla, Richard and Wetter, Thomas and Pfeifer, Daniel},
  journal={Journal of medical Internet research},
  volume={22},
  number={7},
  pages={e17853},
  year={2020},
  publisher={JMIR Publications Toronto, Canada}
}

@inproceedings{peng2019transfer,
  title={Transfer Learning in Biomedical Natural Language Processing: An Evaluation of BERT and ELMo on Ten Benchmarking Datasets},
  author={Peng, Yifan and Yan, Shankai and Lu, Zhiyong},
  booktitle={Proceedings of the 18th BioNLP Workshop and Shared Task},
  pages={58--65},
  year={2019}
}

@inproceedings{beltagy2019scibert,
  title={SciBERT: A Pretrained Language Model for Scientific Text},
  author={Beltagy, Iz and Lo, Kyle and Cohan, Arman},
  booktitle={Proceedings of the 2019 Conference on Empirical Methods in Natural Language Processing and the 9th International Joint Conference on Natural Language Processing (EMNLP-IJCNLP)},
  pages={3615--3620},
  year={2019}
}

@article{huang2019clinicalbert,
  title={Clinicalbert: Modeling clinical notes and predicting hospital readmission},
  author={Huang, Kexin and Altosaar, Jaan and Ranganath, Rajesh},
  journal={arXiv preprint arXiv:1904.05342},
  year={2019}
}

@inproceedings{martin2020camembert,
	address = {Online},
	title = {{CamemBERT}: a {Tasty} {French} {Language} {Model}},
	url = {https://www.aclweb.org/anthology/2020.acl-main.645},
	booktitle = {Proceedings of the 58th {Annual} {Meeting} of the {Association} for {Computational} {Linguistics}},
	publisher = {Association for Computational Linguistics},
	author = {Martin, Louis and Muller, Benjamin and Ortiz Suárez, Pedro Javier and Dupont, Yoann and Romary, Laurent and de la Clergerie, {\'E}ric and Seddah, Djamé and Sagot, Benoît},
	month = jul,
	year = {2020},
	pages = {7203--7219}
}

@inproceedings{chan2020german,
  title={German's Next Language Model},
  author={Chan, Branden and Schweter, Stefan and M{\"o}ller, Timo},
  booktitle={Proceedings of the 28th International Conference on Computational Linguistics},
  pages={6788--6796},
  year={2020}
}

@article{lentzen2022critical,
  title={Critical assessment of transformer-based AI models for German clinical notes},
  author={Lentzen, Manuel and Madan, Sumit and Lage-Rupprecht, Vanessa and K{\"u}hnel, Lisa and Fluck, Juliane and Jacobs, Marc and Mittermaier, Mirja and Witzenrath, Martin and Brunecker, Peter and Hofmann-Apitius, Martin and others},
  journal={JAMIA open},
  volume={5},
  number={4},
  pages={ooac087},
  year={2022},
  publisher={Oxford University Press}
}

@article{arefeva2022tourbert,
  title={When BERT Started Traveling: TourBERT—A Natural Language Processing Model for the Travel Industry},
  author={Arefeva, Veronika and Egger, Roman},
  journal={Digital},
  volume={2},
  number={4},
  pages={546--559},
  year={2022},
  publisher={MDPI}
}

@article{lee2020biobert,
  title={BioBERT: a pre-trained biomedical language representation model for biomedical text mining},
  author={Lee, Jinhyuk and Yoon, Wonjin and Kim, Sungdong and Kim, Donghyeon and Kim, Sunkyu and So, Chan Ho and Kang, Jaewoo},
  journal={Bioinformatics},
  volume={36},
  number={4},
  pages={1234--1240},
  year={2020},
  publisher={Oxford University Press}
}

@inproceedings{ng2019facebook,
  title={Facebook FAIR's WMT19 News Translation Task Submission},
  author={Ng, Nathan and Yee, Kyra and Baevski, Alexei and Ott, Myle and Auli, Michael and Edunov, Sergey},
  booktitle={Proceedings of the Fourth Conference on Machine Translation (Volume 2: Shared Task Papers, Day 1)},
  year={2019},
  organization={Association for Computational Linguistics}
}

@inproceedings{edunov2018understanding,
  title={Understanding Back-Translation at Scale},
  author={Edunov, Sergey and Ott, Myle and Auli, Michael and Grangier, David},
  booktitle={Proceedings of the 2018 Conference on Empirical Methods in Natural Language Processing},
  pages={489--500},
  year={2018}
}

@article{johnson2023mimic,
  title={MIMIC-IV, a freely accessible electronic health record dataset},
  author={Johnson, Alistair EW and Bulgarelli, Lucas and Shen, Lu and Gayles, Alvin and Shammout, Ayad and Horng, Steven and Pollard, Tom J and Hao, Sicheng and Moody, Benjamin and Gow, Brian and others},
  journal={Scientific data},
  volume={10},
  number={1},
  pages={1},
  year={2023},
  publisher={Nature Publishing Group UK London}
}

@misc{johnson2023mimicnote,
  doi = {10.13026/7QGP-KC16},
  url = {https://physionet.org/content/mimic-iv-note/},
  author = {Johnson, Alistair and Pollard, Tom and Horng, Steven and Celi, Leo Anthony and Mark, Roger},
  title = {MIMIC-IV-Note: Deidentified free-text clinical notes},
  publisher = {PhysioNet},
  year = {2023}
}

@article{goldberger2000physiobank,
  title={PhysioBank, PhysioToolkit, and PhysioNet: components of a new research resource for complex physiologic signals},
  author={Goldberger, Ary L and Amaral, Luis AN and Glass, Leon and Hausdorff, Jeffrey M and Ivanov, Plamen Ch and Mark, Roger G and Mietus, Joseph E and Moody, George B and Peng, Chung-Kang and Stanley, H Eugene},
  journal={circulation},
  volume={101},
  number={23},
  pages={e215--e220},
  year={2000},
  publisher={Am Heart Assoc}
}

@inproceedings{mikolov2013distributed,
 author = {Mikolov, Tomas and Sutskever, Ilya and Chen, Kai and Corrado, Greg S and Dean, Jeff},
 booktitle = {Advances in Neural Information Processing Systems},
 pages = {},
 publisher = {Curran Associates, Inc.},
 title = {Distributed Representations of Words and Phrases and their Compositionality},
 volume = {26},
 year = {2013}
}

@article{gojare2015analysis,
  title={Analysis and design of selenium webdriver automation testing framework},
  author={Gojare, Satish and Joshi, Rahul and Gaigaware, Dhanashree},
  journal={Procedia Computer Science},
  volume={50},
  pages={341--346},
  year={2015},
  publisher={Elsevier}
}

@inproceedings{mckinney2010data,
  title={Data structures for statistical computing in python},
  author={McKinney, Wes and others},
  booktitle={Proceedings of the 9th Python in Science Conference},
  volume={445},
  pages={51--56},
  year={2010},
  organization={Austin, TX}
}

@article{smith2006peer,
  title={Peer review: a flawed process at the heart of science and journals},
  author={Smith, Richard},
  journal={Journal of the royal society of medicine},
  volume={99},
  number={4},
  pages={178--182},
  year={2006},
  publisher={SAGE Publications Sage UK: London, England}
}

@inproceedings{vaswani2017attention,
author = {Vaswani, Ashish and Shazeer, Noam and Parmar, Niki and Uszkoreit, Jakob and Jones, Llion and Gomez, Aidan N. and Kaiser, \L{}ukasz and Polosukhin, Illia},
title = {Attention is all you need},
year = {2017},
publisher = {Curran Associates Inc.},
address = {Red Hook, NY, USA},
booktitle = {Proceedings of the 31st International Conference on Neural Information Processing Systems},
pages = {6000–6010},
numpages = {11},
location = {Long Beach, California, USA},
series = {NIPS'17}
}

@inproceedings{wolf2020transformers,
  title={Transformers: State-of-the-art natural language processing},
  author={Wolf, Thomas and Debut, Lysandre and Sanh, Victor and Chaumond, Julien and Delangue, Clement and Moi, Anthony and Cistac, Pierric and Rault, Tim and Louf, R{\'e}mi and Funtowicz, Morgan and others},
  booktitle={Proceedings of the 2020 conference on empirical methods in natural language processing: system demonstrations},
  pages={38--45},
  year={2020}
}

@inproceedings{lhoest2021datasets,
  title={Datasets: A Community Library for Natural Language Processing},
  author={Lhoest, Quentin and del Moral, Albert Villanova and Jernite, Yacine and Thakur, Abhishek and von Platen, Patrick and Patil, Suraj and Chaumond, Julien and Drame, Mariama and Plu, Julien and Tunstall, Lewis and others},
  booktitle={Proceedings of the 2021 Conference on Empirical Methods in Natural Language Processing: System Demonstrations},
  pages={175--184},
  year={2021}
}

@article{achakulvisut2020,
  doi = {10.21105/joss.01979},
  url = {https://doi.org/10.21105/joss.01979},
  year = {2020},
  publisher = {The Open Journal},
  volume = {5},
  number = {46},
  pages = {1979},
  author = {Titipat Achakulvisut and Daniel Acuna and Konrad Kording},
  title = {Pubmed Parser: A Python Parser for PubMed Open-Access XML Subset and MEDLINE XML Dataset XML Dataset},
  journal = {Journal of Open Source Software}
}

@article{khare2004nutch,
  title={Nutch: A flexible and scalable open-source web search engine},
  author={Khare, Rohit and Cutting, Doug and Sitaker, Kragen and Rifkin, Adam},
  journal={Oregon State University},
  volume={1},
  pages={32--32},
  year={2004},
  publisher={Citeseer}
}

@Article{specht2025evaluating,
author="Specht, Lisa
and Scheible, Raphael
and Boeker, Martin
and Farin-Glattacker, Erik
and Kampel, Nikolas
and Schm{\"o}lz, Marina
and Sch{\"o}pf-Lazzarino, Andrea
and Schulz, Stefan
and Schlett, Christian
and Thomczyk, Fabian
and Voigt-Radloff, Sebastian
and Wegner, Constanze
and Wollmann, Katharina
and Maun, Andy",
title="Evaluating the Acceptance and Usability of an Independent, Noncommercial Search Engine for Medical Information: Cross-Sectional Questionnaire Study and User Behavior Tracking Analysis",
journal="JMIR Hum Factors",
year="2025",
month="Jan",
day="23",
volume="12",
pages="e56941",
issn="2292-9495",
doi="10.2196/56941",
url="https://humanfactors.jmir.org/2025/1/e56941",
url="https://doi.org/10.2196/56941"
}

@inproceedings{schabus2017one,
  title={One million posts: A data set of german online discussions},
  author={Schabus, Dietmar and Skowron, Marcin and Trapp, Martin},
  booktitle={Proceedings of the 40th international ACM SIGIR conference on research and development in information retrieval},
  pages={1241--1244},
  year={2017}
}

@article{white2020pubmed,
  title={PubMed 2.0},
  author={White, Jacob},
  journal={Medical reference services quarterly},
  volume={39},
  number={4},
  pages={382--387},
  year={2020},
  publisher={Taylor \& Francis}
}

@article{costa2022no,
  title={No language left behind: Scaling human-centered machine translation},
  author={Costa-juss{\`a}, Marta R and Cross, James and {\c{C}}elebi, Onur and Elbayad, Maha and Heafield, Kenneth and Heffernan, Kevin and Kalbassi, Elahe and Lam, Janice and Licht, Daniel and Maillard, Jean and others},
  journal={arXiv preprint arXiv:2207.04672},
  year={2022}
}

@inproceedings{deng2025crawler,
    booktitle = {Proceedings of MIE 2025},
    year = 2025,
    month = may,
    address = "Glasgow, Scotland",
    publisher = "IOPress",
    title = {Building a Scalable Health Information Crawler: Leveraging Apache Nutch for the tala-med Search Engine},
    language = {en},
    month=may,
    year={2025},
    journal = {Proceedings of MIE 2025},
    
    author = {Deng, Nanxing and Boeker, Martin and Scheible, Raphael},
}

@article{dalianis2009stockholm,
  title={The Stockholm EPR Corpus-characteristics and some initial findings},
  author={Dalianis, Hercules and Hassel, Martin and Velupillai, Sumithra},
  journal={Proceedings of ISHIMR},
  pages={243--249},
  year={2009}
}

@article{wang2018clinical,
  title={Clinical information extraction applications: a literature review},
  author={Wang, Yanshan and Wang, Liwei and Rastegar-Mojarad, Majid and Moon, Sungrim and Shen, Feichen and Afzal, Naveed and Liu, Sijia and Zeng, Yuqun and Mehrabi, Saeed and Sohn, Sunghwan and others},
  journal={Journal of biomedical informatics},
  volume={77},
  pages={34--49},
  year={2018},
  publisher={Elsevier}
}

@article{sager1994natural,
  title={Natural language processing and the representation of clinical data},
  author={Sager, Naomi and Lyman, Margaret and Bucknall, Christine and Nhan, Ngo and Tick, Leo J},
  journal={Journal of the American Medical Informatics Association},
  volume={1},
  number={2},
  pages={142--160},
  year={1994},
  publisher={BMJ Group BMA House, Tavistock Square, London, WC1H 9JR}
}

@inproceedings{borst1991textinfo,
  title={TEXTINFO: a tool for automatic determination of patient clinical profiles using text analysis},
  author={Borst, Francois and Lyman, Margaret and Nhan, NT and Tick, LJ and Sager, N and Scherrer, JR},
  booktitle={Proceedings of the Annual Symposium on Computer Application in Medical Care},
  pages={63},
  year={1991}
}

@inproceedings{friedman1995architectural,
  title={Architectural requirements for a multipurpose natural language processor in the clinical environment},
  author={Friedman, Carol and Johnson, Stephen B and Forman, Bruce and Starren, Justin},
  booktitle={Proceedings of the Annual Symposium on Computer Application in Medical Care},
  pages={347},
  year={1995}
}

@article{zhou2022natural,
  title={Natural language processing for smart healthcare},
  author={Zhou, Binggui and Yang, Guanghua and Shi, Zheng and Ma, Shaodan},
  journal={IEEE Reviews in Biomedical Engineering},
  volume={17},
  pages={4--18},
  year={2022},
  publisher={IEEE}
}

@article{aronson2010overview,
  title={An overview of MetaMap: historical perspective and recent advances},
  author={Aronson, Alan R and Lang, Fran{\c{c}}ois-Michel},
  journal={Journal of the American Medical Informatics Association},
  volume={17},
  number={3},
  pages={229--236},
  year={2010},
  publisher={BMJ Group BMA House, Tavistock Square, London, WC1H 9JR}
}

@article{savova2010mayo,
  title={Mayo clinical Text Analysis and Knowledge Extraction System (cTAKES): architecture, component evaluation and applications},
  author={Savova, Guergana K and Masanz, James J and Ogren, Philip V and Zheng, Jiaping and Sohn, Sunghwan and Kipper-Schuler, Karin C and Chute, Christopher G},
  journal={Journal of the American Medical Informatics Association},
  volume={17},
  number={5},
  pages={507--513},
  year={2010},
  publisher={BMJ Group BMA House, Tavistock Square, London, WC1H 9JR}
}

@inproceedings{friedman2000broad,
  title={A broad-coverage natural language processing system},
  author={Friedman, Carol},
  booktitle={Proceedings of the AMIA Symposium},
  pages={270},
  year={2000}
}

@article{soysal2018clamp,
  title={CLAMP--a toolkit for efficiently building customized clinical natural language processing pipelines},
  author={Soysal, Ergin and Wang, Jingqi and Jiang, Min and Wu, Yonghui and Pakhomov, Serguei and Liu, Hongfang and Xu, Hua},
  journal={Journal of the American Medical Informatics Association},
  volume={25},
  number={3},
  pages={331--336},
  year={2018},
  publisher={Oxford University Press}
}

@article{uzuner20112010,
  title={2010 i2b2/VA challenge on concepts, assertions, and relations in clinical text},
  author={Uzuner, {\"O}zlem and South, Brett R and Shen, Shuying and DuVall, Scott L},
  journal={Journal of the American Medical Informatics Association},
  volume={18},
  number={5},
  pages={552--556},
  year={2011},
  publisher={BMJ Group BMA House, Tavistock Square, London, WC1H 9JR}
}

@article{stubbs2019cohort,
  title={Cohort selection for clinical trials: n2c2 2018 shared task track 1},
  author={Stubbs, Amber and Filannino, Michele and Soysal, Ergin and Henry, Samuel and Uzuner, {\"O}zlem},
  journal={Journal of the American Medical Informatics Association},
  volume={26},
  number={11},
  pages={1163--1171},
  year={2019},
  publisher={Oxford University Press}
}

@article{henry20202018,
  title={2018 n2c2 shared task on adverse drug events and medication extraction in electronic health records},
  author={Henry, Sam and Buchan, Kevin and Filannino, Michele and Stubbs, Amber and Uzuner, Ozlem},
  journal={Journal of the American Medical Informatics Association},
  volume={27},
  number={1},
  pages={3--12},
  year={2020},
  publisher={Oxford University Press}
}

@book{crestani2019experimental,
  title={Experimental IR Meets Multilinguality, Multimodality, and Interaction: 10th International Conference of the CLEF Association, CLEF 2019, Lugano, Switzerland, September 9--12, 2019, Proceedings},
  author={Crestani, Fabio and Braschler, Martin and Savoy, Jacques and Rauber, Andreas and M{\"u}ller, Henning and Losada, David E and B{\"u}rki, Gundula Heinatz and Cappellato, Linda and Ferro, Nicola},
  volume={11696},
  year={2019},
  publisher={Springer Nature}
}

@Misc{clef2019nts,
  author = 	{Neves, Mariana
		and Butzke, Daniel
		and D{\"o}rendahl, Antje
		and Leich, Nora
		and Grune, Barbara
		and Sch{\"o}nfelder, Gilbert},
  title = 	{Non-technical Summaries (NTS) of Animal Experiments Indexed with ICD-10 Codes (Version 1.0)},
  year = 	{2019},
  month = 	{1},
  day = 	{18},
  publisher = 	{Open Agrar Repository},
  doi = 	{10.17590/20190118-134645-0},
  url = 	{https://www.openagrar.de/receive/openagrar_mods_00046540},
  url = 	{https://doi.org/10.17590/20190118-134645-0},
  language = 	{en}
}

@Misc{clef2019test,
  author = 	{Neves, Mariana
		and Butzke, Daniel
		and D{\"o}rendahl, Antje
		and Leich, Nora
		and Grune, Barbara
		and Sch{\"o}nfelder, Gilbert},
  title = 	{Test set of Non-technical Summaries (NTS) of Animal Experiments Indexed with ICD-10 Codes (Version 1.0)},
  year = 	{2019},
  month = 	{5},
  day = 	{06},
  publisher = 	{Open Agrar Repository},
  doi = 	{10.17590/20190506-101759},
  url = 	{https://www.openagrar.de/receive/openagrar_mods_00049062},
  url = 	{https://doi.org/10.17590/20190506-101759},
  language = 	{en}
}

@inproceedings{peters2018dissecting,
  title={Dissecting Contextual Word Embeddings: Architecture and Representation},
  author={Peters, Matthew E and Neumann, Mark and Zettlemoyer, Luke and Yih, Wen-tau},
  booktitle={Proceedings of the 2018 Conference on Empirical Methods in Natural Language Processing},
  pages={1499--1509},
  year={2018}
}

@inproceedings{joulin2017bag,
  title={Bag of Tricks for Efficient Text Classification},
  author={Joulin, Armand and Grave, {\'E}douard and Bojanowski, Piotr and Mikolov, Tom{\'a}{\v{s}}},
  booktitle={Proceedings of the 15th Conference of the European Chapter of the Association for Computational Linguistics: Volume 2, Short Papers},
  pages={427--431},
  year={2017}
}

@inproceedings{akhtyamova2020named,
  title={Named entity recognition in Spanish biomedical literature: Short review and BERT model},
  author={Akhtyamova, Liliya},
  booktitle={2020 26th Conference of Open Innovations Association (FRUCT)},
  pages={1--7},
  year={2020},
  organization={IEEE}
}

@inproceedings{rubel2020biobertpt,
  title={BioBERTpt: a Portuguese neural language model for clinical named entity recognition},
  author={Rubel Schneider, Elisa Terumi and Andrioli de Souza, Jo{\~a}o Vitor and Knafou, Julien and Oliveira, Lucas ES and Gumiel, Yohan B and de Oliveira, Lucas FA and Teodoro, Douglas and Paraiso, Emerson Cabrera and Moro, Claudia and others},
  booktitle={Proceedings of the 3rd Clinical Natural Language Processing Workshop},
  year={2020},
  organization={19 November 2020}
}

@inproceedings{copara2020contextualized,
  title={Contextualized French language models for biomedical named entity recognition},
  author={Copara, Jenny and Knafou, Julien and Naderi, Nona and Moro, Claudia and Ruch, Patrick and Teodoro, Douglas},
  booktitle={6e Conf{\'e}rence Conjointe Journ{\'e}es D'{\'e}tudes Sur La Parole (Jep, 33e Edition), Traitement Automatique Des Langues Naturelles (Taln, 27e Edition), Rencontre Des Etudiants Chercheurs En Informatique Pour Le Traitement Automatique Des Langues (R{\'e}cital, 22e Edition). Atelier D{\'e}fi Fouille De Textes},
  pages={36--48},
  year={2020},
  organization={ATALA; AFCP}
}

@article{starlinger2017improve,
  title={How to improve information extraction from German medical records},
  author={Starlinger, Johannes and Kittner, Madeleine and Blankenstein, Oliver and Leser, Ulf},
  journal={It-Information Technology},
  volume={59},
  number={4},
  pages={171--179},
  year={2017},
  publisher={De Gruyter Oldenbourg}
}

@incollection{hellrich2015sharing,
  title={Sharing models and tools for processing German clinical texts},
  author={Hellrich, Johannes and Matthies, Franz and Faessler, Erik and Hahn, Udo},
  booktitle={Digital Healthcare Empowering Europeans},
  pages={734--738},
  year={2015},
  publisher={IOS Press}
}

@inproceedings{lohr2018sharing,
  author = {Christina Lohr, Sven Buechel and Udo Hahn},
  title = {Sharing Copies of Synthetic Clinical Corpora without Physical Distribution — A Case Study to Get Around IPRs and Privacy Constraints Featuring the German JSYNCC Corpus},
  booktitle = {Proceedings of the Eleventh International Conference on Language Resources and Evaluation (LREC 2018)},
  year = {2018},
  month = may,
  publisher = {European Language Resources Association (ELRA)},
  isbn = {979-10-95546-00-9}
  }

@article{liu2021med,
  title={Med-BERT: A pretraining framework for medical records named entity recognition},
  author={Liu, Ning and Hu, Qian and Xu, Huayun and Xu, Xing and Chen, Mengxin},
  journal={IEEE Transactions on Industrial Informatics},
  volume={18},
  number={8},
  pages={5600--5608},
  year={2021},
  publisher={IEEE}
}

@inproceedings{conneau2020unsupervised,
  title={Unsupervised Cross-lingual Representation Learning at Scale},
  author={Conneau, Alexis and Khandelwal, Kartikay and Goyal, Naman and Chaudhary, Vishrav and Wenzek, Guillaume and Guzm{\'a}n, Francisco and Grave, {\'E}douard and Ott, Myle and Zettlemoyer, Luke and Stoyanov, Veselin},
  booktitle={Proceedings of the 58th Annual Meeting of the Association for Computational Linguistics},
  pages={8440--8451},
  year={2020}
}

@inproceedings{suarez2019asynchronous,
  title={Asynchronous pipeline for processing huge corpora on medium to low resource infrastructures},
  author={Su{\'a}rez, Pedro Javier Ortiz and Sagot, Beno{\^\i}t and Romary, Laurent},
  booktitle={7th Workshop on the Challenges in the Management of Large Corpora (CMLC-7)},
  year={2019},
  organization={Leibniz-Institut f{\"u}r Deutsche Sprache}
}

@article{de2019bertje,
  title={BERTje: A Dutch BERT model},
  author={De Vries, Wietse and van Cranenburgh, Andreas and Bisazza, Arianna and Caselli, Tommaso and van Noord, Gertjan and Nissim, Malvina},
  journal={arXiv preprint arXiv:1912.09582},
  year={2019}
}

@inproceedings{sanger2019classifying,
  title={Classifying German Animal Experiment Summaries with Multi-lingual BERT at CLEF eHealth 2019 Task 1.},
  author={S{\"a}nger, Mario and Weber, Leon and Kittner, Madeleine and Leser, Ulf},
  year={2019}
}

@book{world1992icd,
  title={The ICD-10 classification of mental and behavioural disorders: clinical descriptions and diagnostic guidelines},
  author={World Health Organization},
  volume={1},
  year={1992},
  publisher={World Health Organization}
}

@article{bengio2003neural,
  title={A neural probabilistic language model},
  author={Bengio, Yoshua and Ducharme, R{\'e}jean and Vincent, Pascal and Jauvin, Christian},
  journal={Journal of machine learning research},
  volume={3},
  number={Feb},
  pages={1137--1155},
  year={2003}
}

@misc{cardiode,
    author = {Christoph Dieterich},
    publisher = {heiDATA},
    title = {{CARDIO:DE}},
    year = {2022},
    version = {V5},
    doi = {10.11588/data/AFYQDY},
    url = {https://doi.org/10.11588/data/AFYQDY}
}

@article{fries2022bigbio,
  title={BigBIO: A framework for data-centric biomedical natural language processing},
  author={Fries, Jason and Weber, Leon and Seelam, Natasha and Altay, Gabriel and Datta, Debajyoti and Garda, Samuele and Kang, Sunny and Su, Rosaline and Kusa, Wojciech and Cahyawijaya, Samuel and others},
  journal={Advances in Neural Information Processing Systems},
  volume={35},
  pages={25792--25806},
  year={2022}
}

@inproceedings{harbecke2022only,
  title={Why only Micro-F1? Class Weighting of Measures for Relation Classification},
  author={Harbecke, David and Chen, Yuxuan and Hennig, Leonhard and Alt, Christoph},
  booktitle={Proceedings of NLP Power! The First Workshop on Efficient Benchmarking in NLP},
  pages={32--41},
  year={2022}
}

@inproceedings{tjong2003introduction,
  title={Introduction to the CoNLL-2003 shared task: language-independent named entity recognition},
  author={Tjong Kim Sang, Erik F and De Meulder, Fien},
  booktitle={Proceedings of the seventh conference on Natural language learning at HLT-NAACL 2003-Volume 4},
  pages={142--147},
  year={2003}
}

@inproceedings{dada2023impact,
  title={On the Impact of Cross-Domain Data on German Language Models},
  author={Dada, Amin and Chen, Aokun and Peng, Cheng and Smith, Kaleb E and Idrissi-Yaghir, Ahmad and Seibold, Constantin Marc and Li, Jianning and Heiliger, Lars and Friedrich, Christoph M and Truhn, Daniel and others},
  booktitle={The 2023 Conference on Empirical Methods in Natural Language Processing},
  year={2023}
}

@misc{seqeval,
  title={{seqeval}: A Python framework for sequence labeling evaluation},
  url={https://github.com/chakki-works/seqeval},
  note={Software available from https://github.com/chakki-works/seqeval},
  author={Hiroki Nakayama},
  year={2018},
}

@article{pedregosa2011scikit,
  title={Scikit-learn: Machine learning in Python},
  author={Pedregosa, Fabian and Varoquaux, Ga{\"e}l and Gramfort, Alexandre and Michel, Vincent and Thirion, Bertrand and Grisel, Olivier and Blondel, Mathieu and Prettenhofer, Peter and Weiss, Ron and Dubourg, Vincent and others},
  journal={Journal of machine learning research},
  volume={12},
  number={Oct},
  pages={2825--2830},
  year={2011}
}

@inproceedings{he2021deberta,
title={DeBERTa: Decoding-enhanced BERT with Disentangled Attention},
author={Pengcheng He and Xiaodong Liu and Jianfeng Gao and Weizhu Chen},
booktitle={International Conference on Learning Representations},
year={2021}
}

@article{nguyen2023culturax,
  title={Culturax: A cleaned, enormous, and multilingual dataset for large language models in 167 languages},
  author={Nguyen, Thuat and Van Nguyen, Chien and Lai, Viet Dac and Man, Hieu and Ngo, Nghia Trung and Dernoncourt, Franck and Rossi, Ryan A and Nguyen, Thien Huu},
  journal={arXiv preprint arXiv:2309.09400},
  year={2023}
}

@inproceedings{tiedemann2012parallel,
  title={Parallel Data, Tools and Interfaces in OPUS},
  author={Tiedemann, J{\"o}rg},
  booktitle={Proceedings of the Eighth International Conference on Language Resources and Evaluation (LREC'12)},
  pages={2214--2218},
  year={2012}
}

@article{beltagy2020longformer,
  title={Longformer: The long-document transformer},
  author={Beltagy, Iz and Peters, Matthew E and Cohan, Arman},
  journal={arXiv preprint arXiv:2004.05150},
  year={2020}
}

@inproceedings{xiong2021nystromformer,
  title={Nystr{\"o}mformer: A nystr{\"o}m-based algorithm for approximating self-attention},
  author={Xiong, Yunyang and Zeng, Zhanpeng and Chakraborty, Rudrasis and Tan, Mingxing and Fung, Glenn and Li, Yin and Singh, Vikas},
  booktitle={Proceedings of the AAAI conference on artificial intelligence},
  volume={35},
  number={16},
  pages={14138--14148},
  year={2021}
}

@misc{shalumov2023herorobertalongformerhebrew,
      title={HeRo: RoBERTa and Longformer Hebrew Language Models}, 
      author={Vitaly Shalumov and Harel Haskey},
      year={2023},
      eprint={2304.11077},
      archivePrefix={arXiv},
      primaryClass={cs.CL},
      url={https://arxiv.org/abs/2304.11077}, 
}

@article{warner2024smarter,
  title={Smarter, better, faster, longer: A modern bidirectional encoder for fast, memory efficient, and long context finetuning and inference},
  author={Warner, Benjamin and Chaffin, Antoine and Clavi{\'e}, Benjamin and Weller, Orion and Hallstr{\"o}m, Oskar and Taghadouini, Said and Gallagher, Alexis and Biswas, Raja and Ladhak, Faisal and Aarsen, Tom and others},
  journal={arXiv preprint arXiv:2412.13663},
  year={2024}
}

@inproceedings{gururangan2020don,
  title={Don’t Stop Pretraining: Adapt Language Models to Domains and Tasks},
  author={Gururangan, Suchin and Marasovi{\'c}, Ana and Swayamdipta, Swabha and Lo, Kyle and Beltagy, Iz and Downey, Doug and Smith, Noah A},
  booktitle={Proceedings of the 58th Annual Meeting of the Association for Computational Linguistics},
  pages={8342--8360},
  year={2020}
}

@inproceedings{el2022re,
  title={Re-train or Train from Scratch? Comparing Pre-training Strategies of BERT in the Medical Domain},
  author={El Boukkouri, Hicham and Ferret, Olivier and Lavergne, Thomas and Zweigenbaum, Pierre},
  booktitle={Proceedings of the Thirteenth Language Resources and Evaluation Conference},
  pages={2626--2633},
  year={2022}
}

@incollection{meineke2023announcement,
  title={Announcement of the German medical text corpus project (GeMTeX)},
  author={Meineke, Frank and Modersohn, Luise and Loeffler, Markus and Boeker, Martin},
  booktitle={Caring is Sharing--Exploiting the Value in Data for Health and Innovation},
  pages={835--836},
  year={2023},
  publisher={IOS Press}
}

@article{tayefi2021challenges,
  title={Challenges and opportunities beyond structured data in analysis of electronic health records},
  author={Tayefi, Maryam and Ngo, Phuong and Chomutare, Taridzo and Dalianis, Hercules and Salvi, Elisa and Budrionis, Andrius and Godtliebsen, Fred},
  journal={Wiley Interdisciplinary Reviews: Computational Statistics},
  volume={13},
  number={6},
  pages={e1549},
  year={2021},
  publisher={Wiley Online Library}
}

@inproceedings{scheibleschmitt2025geistbertbreathinglifegerman,
    title = "{G}eist{BERT}: Breathing Life into {G}erman {NLP}",
    author = "Scheible-Schmitt, Raphael  and
      Frei, Johann",
    editor = "Das, Sudhansu Bala  and
      Mishra, Pruthwik  and
      Singh, Alok  and
      Muhammad, Shamsuddeen Hassan  and
      Ekbal, Asif  and
      Das, Uday Kumar",
    booktitle = "Proceedings of the Workshop on Beyond English: Natural Language Processing for all Languages in an Era of Large Language Models",
    month = sep,
    year = "2025",
    address = "Varna, Bulgaria",
    publisher = "INCOMA Ltd., Shoumen, BULGARIA",
    url = "https://aclanthology.org/2025.globalnlp-1.6/",
    pages = "42--50"
}

@misc{springernature,
  author = {{Springer Nature}},
  title = {Springer Nature Developer Portal – APIs for Research Papers},
  url = {https://dev.springernature.com/},
  note = {Accessed on 2024-11-10},
  year = {2024}
}

@misc{wikipediaexport,
  author = {{Wikipedia contributors}},
  title = {Seiten exportieren – Wikipedia},
  url = {https://de.wikipedia.org/wiki/Spezial:Exportieren},
  note = {Accessed on 2024-11-12},
  year = {2024}
}

@misc{pmcoa,
  author = {{National Library of Medicine}},
  title = {PMC Open Access Subset},
  url = {https://pmc.ncbi.nlm.nih.gov/tools/openftlist/},
  note = {Accessed on 2025-01-12},
  year = {2003}
}

@misc{refubium,
  author = {{Freie Universität Berlin}},
  title = {Refubium – Home},
  url = {https://refubium.fu-berlin.de/},
  note = {Accessed on 2024-11-10},
  year = {2024}
}

@misc{10kGNAD,
  author = {tblock},
  title = {10kGNAD},
  url = {https://tblock.github.io/10kGNAD/},
  note = {Accessed on 2025-02-05},
  year = {2025}
}

@misc{clean-text,
  author = {jfilter},
  title = {clean-text},
  url = {https://github.com/jfilter/clean-text},
  note = {Accessed on 2025-02-05},
  year = {2025}
}

@misc{labelstudio,
  author = {Maxim Tkachenko and Mikhail Malyuk and Andrey Holmanyuk and Nikolai Liubimov},
  title = {Label Studio: Data Labeling Software},
  url = {https://github.com/heartexlabs/label-studio},
  note = {Accessed on 2025-02-05},
  year = {2022}
}

@misc{nllbAPI,
  author = {winstxnhdw},
  title = {nllb-api},
  url = {https://github.com/winstxnhdw/nllb-api},
  note = {Accessed on 2025-02-15},
  year = {2025}
}

@misc{animaltestinfo,
  author = {{German Federal Institute for Risk Assessment}},
  title = {AnimalTestInfo},
  url = {https://animaltestinfo.de/},
  note = {Accessed on 2025-04-28},
  year = {2025}
}

@misc{nni,
  author = {{Microsoft}},
  title = {Neural Network Intelligence},
  url = {https://github.com/microsoft/nni/},
  note = {Accessed on 2025-05-01},
  year = {2025}
}

@misc{kurtovic_earwigmwparserfromhell_2025,
	title = {earwig/mwparserfromhell},
	copyright = {MIT},
	url = {https://github.com/earwig/mwparserfromhell},
	urldate = {2025-07-31},
	author = {Kurtovic, Ben},
	month = jul,
	year = {2025},
	note = {original-date: 2012-05-20T18:45:54Z},
}

@misc{christbertscignad_2024,
  title = {{ChristBERT}/{sciGNAD} · {Datasets} at {Hugging} {Face}},
  url = {https://huggingface.co/datasets/ChristBERT/sciGNAD},
  author = {He, Henry},
  year = {2024},
  note = {Accessed on 2025-07-31}
}

@misc{christbertscignad_tcls_2024,
  title = {{ChristBERT}/{sciGNAD}\_tcls · {Hugging} {Face}},
  url = {https://huggingface.co/ChristBERT/sciGNAD_tcls},
  author = {Schmitt, Raphael},
  year = {2024},
  note = {Accessed on 2025-07-31}
}

\newpage
\appendix


\graphicspath{ {./figures/} }

\setcounter{table}{0}
\setcounter{figure}{0}
\setcounter{equation}{0}
\renewcommand{\thetable}{S\arabic{table}}
\renewcommand{\thefigure}{S\arabic{figure}}
\renewcommand{\theequation}{S\arabic{equation}}





\begin{center}
{\LARGE Supplementary Material}
\end{center}

\vspace{1cm}

\section{Perplexity}

\begin{figure}[H]
  \centering
  \includegraphics[width=0.95\textwidth]{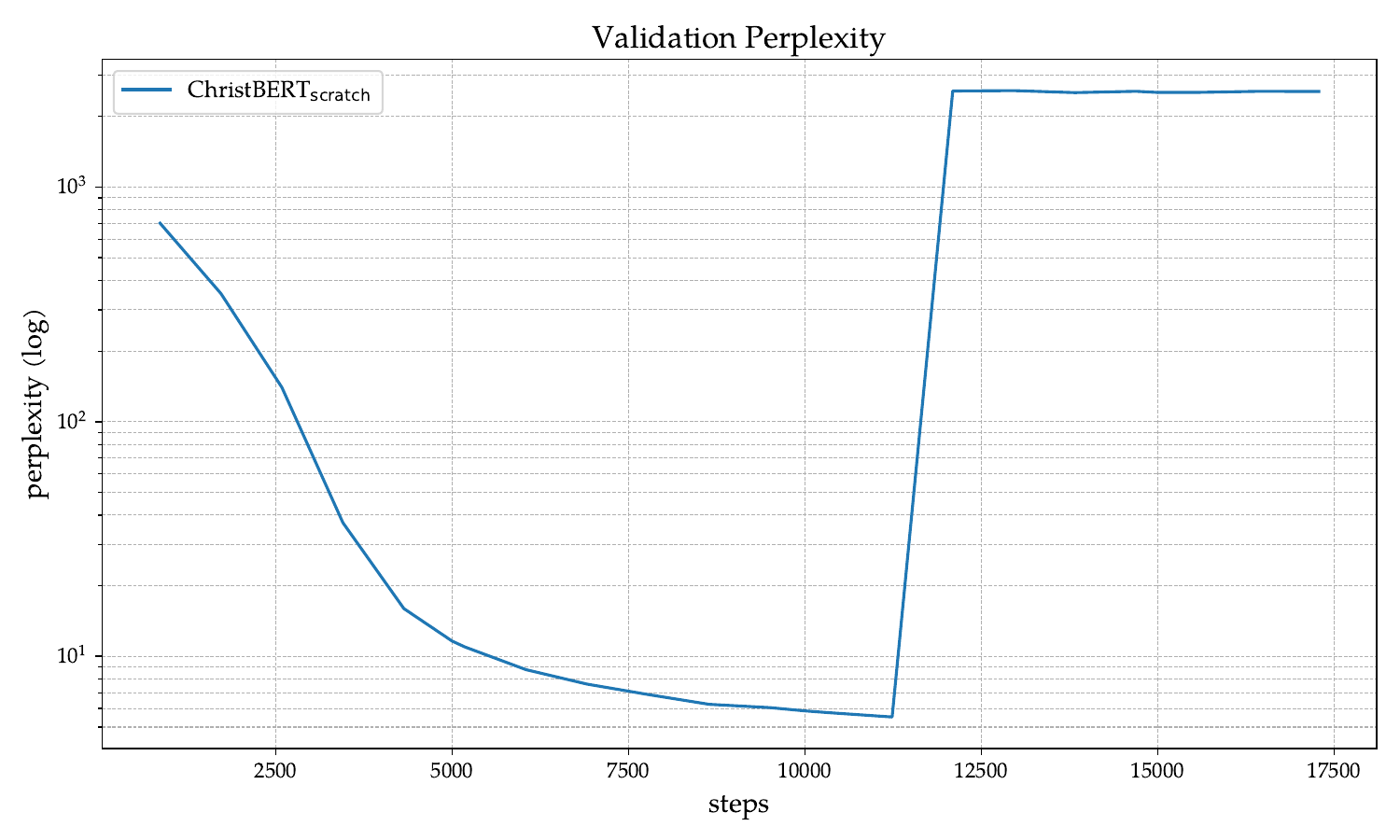}
  \caption[Perplexity during diverged pre-training of
  \ChristBERT\textsubscript{scratch}]{Perplexity during diverged pre-training of
  \ChristBERT\textsubscript{scratch}. Perplexity is shown in log scale for every
  optimization step and evaluated on the validation split of the pre-training
  corpus. The plot illustrates a sharp increase in perplexity around the
  12,500th step, indicating model instability and failure to converge.}
  \label{fig:div_ppl}
\end{figure}

Figure~\ref{fig:div_ppl} illustrates the training instability observed during the diverged pre-training of \ChristBERT\textsubscript{scratch}. The plot shows perplexity on the validation split of the pre-training corpus across optimization steps. A sharp increase in perplexity is visible around step 12,500, indicating a failure to converge.

\section{Model Properties}

Table~\ref{tab:model_props} summarizes the vocabulary size and number of parameters for each evaluated model. While this table focuses on model size, other architectural differences are not shown.

\begin{figure}[H]
\captionsetup{type=table}
  \centering 
  \begin{tabular}{l cc}
    \toprule
    \bfseries Model & \bfseries Vocab Size & \bfseries \# Parameters \\ 
    \midrule
    \ChristBERT & 52,009 & 125,985,024 \\
    \ChristBERT\textsubscript{scratch} & 52,009 & 125,985,024 \\
    \ChristBERT\textsubscript{BPE} & 52,009 & 125,985,024 \\
    medBERT.de & 30,000 & 109,081,344 \\
    BioGottBERT & 52,009 & 125,985,024 \\
    GeistBERT & 52,009 & 125,985,024 \\
    GeBERTa & 50,266 & 138,620,928 \\
    \bottomrule
\end{tabular}

  \caption[Vocabulary size and parameter size of evaluated models]{The
  vocabulary size and parameter size are shown for the evaluated models. This
  table does not show other design differences of the models. Values extracted
  using \textsc{Huggingface Transformers} library.}
  \label{tab:model_props}
\end{figure}

\section{Timing and Hyperparameter Search Overview}

The total computation time required for pre-training is detailed in Table~\ref{tab:train_gpu_time}. In addition, Table~\ref{tab:eval_total_time} reports the time spent on hyperparameter grid search for all downstream tasks, performed on a single NVIDIA RTX 3090 GPU. Table~\ref{tab:eval_gpu_time} lists the fine-tuning (FT) and inference (PT) runtimes for the final selected models, also measured on the same hardware.

The best-performing hyperparameter configurations (batch size and learning rate) for each task and model are provided in Table~\ref{tab:best_hyperparams}.

\begin{table}[htb]
    \centering
    \begin{tabular}{lclcc}
    \toprule
    \multirow{2}{*}{\bfseries Model} &
    \bfseries Computation Time &
    \multicolumn{2}{c}{\multirow{2}{*}{\bfseries GPUs}} &
    \multirow{2}{*}{\bfseries VRAM} \\
    & (DD:HH:MM) & & & \\
    \midrule
    \ChristBERT{} & 6:20:09 & 4 $\times$ A100 SXM && 80 GB \\
    \ChristBERT\textsubscript{scratch} & 6:19:13 & 4 $\times$ A100 SXM && 80 GB \\
    \ChristBERT\textsubscript{BPE} & 7:09:12 & 2 $\times$ H100 && 93 GB \\
    \bottomrule
\end{tabular}

    \caption[Computation time of pre-training]{Pre-training computation time in
    days, hours and minutes summing up to 521 hours and 54 minutes, which are
    approximately 21.74 days.}
    \label{tab:train_gpu_time}
\end{table}

\begin{table}[htbp]
    \centering

\begin{tabular}{l ccccc}
    \toprule
    \bfseries Model & \bfseries BRONCO150 & \bfseries CARDIO:DE & 
    \bfseries GGPONC & \bfseries CLEF & \bfseries JSynCC \\
    \midrule
    \ChristBERT & 1:20:26 & 4:40:01 & 11:26:14 & 11:24:22 & 2:10:02 \\
    \ChristBERT\textsubscript{scratch} & 1:28:28 & 5:14:12 & 10:47:27 & 12:16:22 & 2:30:59 \\
    \ChristBERT\textsubscript{BPE} & 1:12:09 & 4:55:18 & 10:51:00 & 11:45:43 & 2:25:13 \\
    medBERT.de & 1:57:32 & 4:31:53 & 11:25:21 & 12:53:27 & 2:13:56 \\
    BioGottBERT & 1:18:15 & 6:01:10 & 10:55:55 & 12:50:17 & 2:03:28 \\
    GeistBERT & 1:27:57 & 4:40:46 & 11:25:43 & 12:51:41 & 2:15:12 \\
    GeBERTa & 1:57:32 & 7:39:51 & 19:22:22 & 29:15:23 & 3:09:10 \\
    \bottomrule
\end{tabular}

    \caption[Computation time of hyperparameter grid search]{Computation time in
    hours, minutes and seconds spent on the hyperparameter grid search for
    finding the best models for each task. The grid search was performed on a
    single NVIDIA RTX 3090 GPU with 24 GB VRAM. The total computation time for
    hyperparameter optimization sums up to 161 hours and 46 minutes, which are
    approximately 6.74 days.}
    \label{tab:eval_total_time}
\end{table}

\begin{table}[htbp]
    \centerline{ 

\begin{tabular}{l cc cc cc cc cc}
    \toprule
    \multirow{2}{*}[-0.5\dimexpr \aboverulesep + \belowrulesep + \cmidrulewidth]{\bfseries Model} &
    \multicolumn{2}{c}{\bfseries BRONCO150} &
    \multicolumn{2}{c}{\bfseries CARDIO:DE} &
    \multicolumn{2}{c}{\bfseries GGPONC} &
    \multicolumn{2}{c}{\bfseries CLEF} &
    \multicolumn{2}{c}{\bfseries JSynCC} \\
    \cmidrule(lr){2-3} \cmidrule(lr){4-5} \cmidrule(lr){6-7} \cmidrule(lr){8-9} \cmidrule(lr){10-11}
    & FT & PT & FT & PT & FT & PT & FT & PT & FT & PT \\
    \midrule
    \ChristBERT & 03:51 & 0.13 & 18:27 & 1.30 & 30:06 & 5.08 & 26:33 & 0.91 & 03:05 & 0.18 \\
    \ChristBERT\textsubscript{scratch} & 02:40 & 0.11 & 15:16 & 0.87 & 30:03 & 5.14 & 29:51 & 0.94 & 06:19 & 0.18 \\
    \ChristBERT\textsubscript{BPE} & 02:03 & 0.13 & 11:09 & 1.01 & 24:53 & 5.45 & 26:32 & 1.08 & 07:16 & 0.39 \\
    medBERT.de & 04:17 & 0.13 & 08:07 & 0.95 & 25:32 & 5.81 & 27:30 & 1.06 & 04:56 & 0.21 \\
    BioGottBERT & 05:46 & 0.11 & 16:48 & 0.92 & 32:47 & 5.14 & 29:13 & 1.05 & 05:28 & 0.19 \\
    GeistBERT & 04:05 & 0.19 & 15:14 & 0.88 & 29:57 & 4.83 & 29:45 & 0.96 & 04:09 & 0.19 \\
    GeBERTa & 04:17 & 0.13 & 28:47 & 1.36 & 49:41 & 7.23 & 56:48 & 2.05 & 06:11 & 0.41 \\
    \bottomrule
\end{tabular}
} 
    \caption[Computation time of fine-tuning and inference]{Fine-tuning (FT)
    runtime in minutes and seconds, and prediction runtime (PT) in seconds of
    the best downstream task models for each task. Both were performed on one
    NVIDIA RTX 3090 GPU with 24 GB VRAM.}
    \label{tab:eval_gpu_time}
\end{table}

\begin{table}[htbp]
    \centerline{

\begin{tabular}{l cc cc cc cc cc}
    \toprule
    \multirow{2}{*}[-0.5\dimexpr \aboverulesep + \belowrulesep + \cmidrulewidth]{\bfseries Model} & 
    \multicolumn{2}{c}{\bfseries BRONCO150} &
    \multicolumn{2}{c}{\bfseries CARDIO:DE} &
    \multicolumn{2}{c}{\bfseries GGPONC} &
    \multicolumn{2}{c}{\bfseries CLEF} &
    \multicolumn{2}{c}{\bfseries JSynCC} \\
    \cmidrule(lr){2-3} \cmidrule(lr){4-5} \cmidrule(lr){6-7} \cmidrule(lr){8-9} \cmidrule(lr){10-11}
    & BS & LR & BS & LR & BS & LR & BS & LR & BS & LR \\
    \midrule
    \ChristBERT & 48 & \num{7e-5} & 48 & \num{7e-5} & 16 & \num{7e-5} & 16 & \num{5e-5} & 48 & \num{5e-5} \\
    \ChristBERT\textsubscript{scratch} & 32 & \num{5e-5} & 16 & \num{5e-5} & 16 & \num{7e-5} & 16 & \num{2e-5} & 64 & \num{5e-5} \\
    \ChristBERT\textsubscript{BPE} & 32 & \num{7e-5} & 32 & \num{5e-5} & 32 & \num{7e-5} & 16 & \num{7e-5} & 16 & \num{5e-6} \\
    medBERT.de & 16 & \num{5e-5} & 48 & \num{7e-5} & 32 & \num{5e-5} & 32 & \num{7e-5} & 64 & \num{2e-5} \\
    BioGottBERT & 16 & \num{7e-5} & 16 & \num{5e-5} & 16 & \num{7e-5} & 16 & \num{7e-5} & 16 & \num{7e-5} \\
    GeistBERT & 16 & \num{2e-5} & 16 & \num{5e-5} & 16 & \num{5e-5} & 16 & \num{2e-5} & 16 & \num{7e-5} \\
    GeBERTa & 16 & \num{5e-5} & 16 & \num{7e-5} & 16 & \num{5e-5} & 48 & \num{7e-5} & 32 & \num{5e-5} \\
    \bottomrule
\end{tabular}
}
    \caption[Best hyperparameters found in the grid search for the downstream
    tasks]{Hyperparameters of the best downstream task models for each task and
    pre-trained model. BS and LR denote batch size and learning rate,
    respectively.}
    \label{tab:best_hyperparams}
\end{table}

\section{Downstream Task Evaluation}

Tables~\ref{tab:bronco} through~\ref{tab:jsyncc} present a detailed breakdown of evaluation results on the downstream tasks. For each dataset, BRONCO150 (Table~\ref{tab:bronco}), CARDIO:DE (Table~\ref{tab:cardiode}), GGPONC (Table~\ref{tab:ggponc2}), and JSynCC (Table~\ref{tab:jsyncc}), the precision, recall, and F1-scores are reported for each class or entity.

All results are shown as percentages and refer to the best fine-tuned model selected based on validation set performance out of 28 grid search runs. The best results are highlighted in bold and the second-best are underlined.

\begin{table}[htbp]
  \centering

\begin{tabular}{l ccc ccc ccc}
    \toprule
    \multirow{2}{*}[-0.5\dimexpr \aboverulesep + \belowrulesep + \cmidrulewidth]{\bfseries Model} &
    \multicolumn{3}{c}{\bfseries Diagnosis} &
    \multicolumn{3}{c}{\bfseries Medication} &
    \multicolumn{3}{c}{\bfseries Treatment} \\
    \cmidrule(lr){2-4} \cmidrule(lr){5-7} \cmidrule(lr){8-10}
    & Prec. & Rec. & \ff & 
    Prec. & Rec. & \ff & 
    Prec. & Rec. & \ff \\
    \midrule
    \ChristBERT & \underline{79.78} & \textbf{81.56} & \underline{80.66} & 85.71 & 81.36 & 83.48 & 81.89 & 83.87 & 82.87 \\
    \ChristBERT\textsubscript{scratch} & 78.89 & 79.33 & 79.11 & \underline{87.50} & 83.05 & \underline{85.22} & \underline{83.59} & \underline{86.29} & \underline{84.92} \\
    \ChristBERT\textsubscript{BPE} & \textbf{82.63} & \underline{80.00} & \textbf{81.29} & \textbf{88.41} & 84.72 & \textbf{86.52} & \textbf{88.82} & \textbf{88.82} & \textbf{88.82} \\
    medBERT.de & 75.35 & 75.35 & 75.35 & 85.71 & 83.33 & 84.51 & 80.13 & 84.03 & 82.03 \\
    BioGottBERT & 72.07 & 72.07 & 72.07 & 83.33 & \underline{84.75} & 84.03 & 80.77 & 84.68 & 82.68 \\
    GeistBERT & 74.05 & 76.54 & 75.27 & 81.25 & \textbf{88.14} & 84.55 & 75.19 & 80.65 & 77.82 \\
    GeBERTa & 75.35 & 75.35 & 75.35 & 85.71 & 83.33 & 84.51 & 80.13 & 84.03 & 82.03 \\
    \bottomrule
\end{tabular}
  
  \caption[Overview of per entity precision, recall and \ff{} scores achieved on
  the BRONCO150 dataset]{Overview of per entity precision (Prec.), recall (Rec.)
  and \ff{} scores achieved on the BRONCO150 dataset All results are shown in
  percent and assess each model's best fine-tuned performance on the test set.
  The best model was selected out of 28 runs based on its validation set
  performance. Best score in bold and second best underlined.}
  \label{tab:bronco}
\end{table}

\begin{table}[htbp]
  \centering

\begin{tabular}{l ccc ccc ccc}
    \toprule
    \multirow{2}{*}[-0.5\dimexpr \aboverulesep + \belowrulesep + \cmidrulewidth]{\bfseries Model} &
    \multicolumn{3}{c}{\bfseries ActiveIng} &
    \multicolumn{3}{c}{\bfseries Drug} &
    \multicolumn{3}{c}{\bfseries Duration} \\
    \cmidrule(lr){2-4} \cmidrule(lr){5-7} \cmidrule(lr){8-10}
    & Prec. & Rec. & \ff &
    Prec. & Rec. & \ff &
    Prec. & Rec. & \ff \\
    \midrule
    \ChristBERT & 85.71 & \textbf{92.96} & \underline{89.19} & 84.85 & 86.15 & 85.50 & 50.00 & \underline{60.00} & 54.55 \\
    \ChristBERT\textsubscript{scratch} & 85.62 & 90.14 & 87.82 & 84.38 & 83.08 & 83.72 & \textbf{59.26} & 58.18 & \underline{58.72} \\
    \ChristBERT\textsubscript{BPE} & \textbf{88.52} & \underline{92.40} & \textbf{90.41} & \underline{91.14} & \textbf{91.14} & \textbf{91.14} & \underline{58.82} & \textbf{60.61} & \textbf{59.70} \\
    medBERT.de & 85.93 & 90.06 & 87.95 & 88.89 & 88.89 & \underline{88.89} & 46.97 & 49.21 & 48.06 \\
    BioGottBERT & 86.29 & 90.85 & 88.51 & 87.30 & 84.62 & 85.94 & 50.85 & 54.55 & 52.63 \\
    GeistBERT & 84.49 & 90.14 & 87.22 & 79.17 & 87.69 & 83.21 & 45.59 & 56.36 & 50.41 \\
    GeBERTa & \underline{88.24} & 89.55 & 88.89 & \textbf{92.31} & \underline{90.00} & \textbf{91.14} & 55.17 & 50.79 & 52.89 \\
    
    \midrule
    & \multicolumn{3}{c}{\bfseries Form} &
    \multicolumn{3}{c}{\bfseries Frequency} &
    \multicolumn{3}{c}{\bfseries Strength} \\
    \cmidrule(lr){2-4} \cmidrule(lr){5-7} \cmidrule(lr){8-10}
    \ChristBERT & 20.00 & \underline{25.00} & 22.22 & 94.63 & \textbf{96.04} & 95.33 & \underline{97.10} & 95.26 & 96.17 \\
    \ChristBERT\textsubscript{scratch} & \textbf{50.00} & \textbf{50.00} & \textbf{50.00} & 93.10 & 93.56 & 93.33 & \textbf{97.16} & \underline{97.16} & \textbf{97.16} \\
    \ChristBERT\textsubscript{BPE} & 16.67 & \underline{25.00} & 20.00 & 95.06 & 94.29 & 94.67 & 94.94 & 96.06 & 95.50 \\
    medBERT.de & \textbf{50.00} & \textbf{50.00} & \textbf{50.00} & 95.00 & \underline{95.87} & 95.43 & 95.15 & 96.86 & 96.00 \\
    BioGottBERT & \textbf{50.00} & \textbf{50.00} & \textbf{50.00} & \underline{96.04} & \textbf{96.04} & \textbf{96.04} & 95.37 & \textbf{97.63} & \underline{96.49} \\
    GeistBERT & \underline{33.33} & \textbf{50.00} & \underline{40.00} & 93.60 & 94.06 & 93.83 & 96.15 & 94.79 & 95.47 \\
    GeBERTa & \textbf{50.00} & \textbf{50.00} & \textbf{50.00} & \textbf{96.37} & 95.60 & \underline{95.98} & 96.03 & 96.41 & 96.22 \\
    \bottomrule
\end{tabular}
  
  \caption[Overview of per entity precision, recall and \ff{} scores achieved on
  the CARDIO:DE dataset]{Overview of per entity precision (Prec.), recall (Rec.)
  and \ff{} scores achieved on the CARDIO:DE dataset All results are shown in
  percent and assess each model's best fine-tuned performance on the test set.
  The best model was selected out of 28 runs based on its validation set
  performance. Best score in bold and second best underlined.}
  \label{tab:cardiode}
\end{table}

\begin{table}[htbp]
  \centering
  \begin{tabular}{l ccc ccc ccc}
    \toprule
    \multirow{3}{*}[-0.5\dimexpr \aboverulesep + \belowrulesep + \cmidrulewidth]{\bfseries Model} &
    \multicolumn{3}{c}{\multirow{2}{*}{\bfseries Clinical}} &
    \multicolumn{3}{c}{\bfseries Diagnosis /} &
    \multicolumn{3}{c}{\multirow{2}{*}{\bfseries Diagnostic}} \\
    & & & & \multicolumn{3}{c}{\bfseries {Pathology}} & & & \\
    \cmidrule(lr){2-4} \cmidrule(lr){5-7} \cmidrule(lr){8-10}
    & Prec. & Rec. & \ff & Prec. & Rec. & \ff & Prec. & Rec. & \ff \\
    \midrule
    \ChristBERT & 79.12 & \textbf{84.28} & \underline{81.62} & 80.26 & 80.81 & 80.53 & 73.34 & 76.18 & 74.73 \\
    \ChristBERT\textsubscript{scratch} & \underline{79.87} & \underline{83.60} & \textbf{81.69} & 80.35 & \textbf{81.67} & \textbf{81.01} & \textbf{73.78} & 76.82 & \underline{75.27} \\
    \ChristBERT\textsubscript{BPE} & \textbf{80.14} & 82.86 & 81.48 & \textbf{80.66} & \underline{81.31} & \underline{80.98} & \underline{73.62} & \textbf{77.54} & \textbf{75.53} \\
    medBERT.de & 76.29 & 81.02 & 78.58 & 78.46 & 78.88 & 78.67 & 72.09 & 74.19 & 73.13 \\
    BioGottBERT & 79.14 & 80.99 & 80.05 & 78.47 & 79.82 & 79.14 & 73.15 & 74.46 & 73.80 \\
    GeistBERT & 79.57 & 81.26 & 80.41 & 78.55 & 79.25 & 78.90 & 72.21 & 75.26 & 73.70 \\
    GeBERTa & 79.81 & 83.08 & 81.42 & \underline{80.39} & 81.22 & 80.80 & 72.93 & \underline{77.09} & 74.95 \\

    \midrule
    & \multicolumn{3}{c}{\multirow{2}{*}{\bfseries External Substance}} &
    \multicolumn{3}{c}{\bfseries Nutrient /} &
    \multicolumn{3}{c}{\multirow{2}{*}{\bfseries Other Finding}} \\
    & & & & \multicolumn{3}{c}{\bfseries Body Substance} & & & \\
    \cmidrule(lr){2-4} \cmidrule(lr){5-7} \cmidrule(lr){8-10}
    \ChristBERT & 56.47 & \underline{53.93} & \underline{55.17} & \textbf{76.11} & \underline{72.11} & \textbf{74.05} & 67.35 & 67.01 & 67.18 \\
    \ChristBERT\textsubscript{scratch} & 50.54 & 52.81 & 51.65 & \underline{73.90} & 70.79 & \underline{72.31} & \textbf{68.78} & \textbf{67.88} & \textbf{68.33} \\
    \ChristBERT\textsubscript{BPE} & 57.43 & \textbf{58.00} & \textbf{57.71} & 70.74 & 71.43 & 71.08 & \underline{68.45} & \underline{67.71} & \underline{68.08} \\
    medBERT.de & 52.87 & 52.27 & 52.57 & 65.48 & 69.25 & 67.31 & 64.50 & 64.56 & 64.53 \\
    BioGottBERT & \underline{58.90} & 48.31 & 53.09 & 71.35 & 69.47 & 70.40 & 67.15 & 64.68 & 65.89 \\
    GeistBERT & 55.42 & 51.69 & 53.49 & 69.17 & \textbf{72.63} & 70.86 & 65.17 & 64.36 & 64.77 \\
    GeBERTa & \textbf{59.49} & 51.09 & 54.97 & 73.42 & 70.28 & 71.81 & 66.85 & 67.12 & 66.98 \\

    \midrule
    & \multicolumn{3}{c}{\bfseries Therapeutic} &
    \multicolumn{3}{c}{} &
    \multicolumn{3}{c}{} \\
    \cmidrule(lr){2-4}
    \ChristBERT & \textbf{79.55} & 79.93 & \textbf{79.74} & & & & & & \\
    \ChristBERT\textsubscript{scratch} & 79.06 & \textbf{80.18} & 79.62 & & & & & & \\
    \ChristBERT\textsubscript{BPE} & \underline{79.41} & \underline{80.00} & \underline{79.70} & & & & & & \\
    medBERT.de & 77.09 & 77.41 & 77.25 & & & & & & \\
    BioGottBERT & 78.01 & 78.63 & 78.32 & & & & & & \\
    GeistBERT & 77.71 & 78.73 & 78.22 & & & & & & \\
    GeBERTa & 78.85 & 79.03 & 78.94 & & & & & & \\
    \bottomrule
\end{tabular}
  
  \caption[Overview of per entity precision, recall and \ff{} scores achieved on
  the GGPONC dataset]{Overview of per entity precision (Prec.), recall (Rec.)
  and \ff{} scores achieved on the GGPONC dataset All results are shown in percent
  and assess each model's best fine-tuned performance on the test set. The best
  model was selected out of 28 runs based on its validation set performance.
  Best score in bold and second best underlined.}
  \label{tab:ggponc2}
\end{table}

\begin{table}[htbp]
  \centering

\begin{tabular}{l ccc ccc ccc}
    \toprule
    \multirow{3}{*}[-0.5\dimexpr \aboverulesep + \belowrulesep + \cmidrulewidth]{\bfseries Model} &
    \multicolumn{3}{c}{\bfseries Trauma} &
    \multicolumn{3}{c}{\multirow{2}{*}{\bfseries Ophthalmology}} &
    \multicolumn{3}{c}{\multirow{2}{*}{\bfseries Orthopedics}} \\
    & \multicolumn{3}{c}{\bfseries Surgery} & & & \\
    \cmidrule(lr){2-4} \cmidrule(lr){5-7} \cmidrule(lr){8-10}
    & Prec. & Rec. & \ff & 
    Prec. & Rec. & \ff & 
    Prec. & Rec. & \ff \\
    \midrule
    \ChristBERT & 84.21 & \textbf{100} & \textbf{91.43} & 100 & 100 & 100 & 89.19 & \textbf{100} & 94.29 \\
    \ChristBERT\textsubscript{scratch} & 83.78 & \underline{96.88} & 89.86 & 100 & 100 & 100 & \textbf{96.97} & \underline{96.97} & \textbf{96.97} \\
    \ChristBERT\textsubscript{BPE} & 82.35 & 87.50 & 84.85 & 100 & 100 & 100 & 91.67 & \textbf{100} & \underline{95.65} \\
    medBERT.de & 83.87 & 81.25 & 82.54 & 100 & 100 & 100 & \underline{93.94} & 93.94 & 93.94 \\
    BioGottBERT & 84.21 & \textbf{100} & \textbf{91.43} & 100 & 100 & 100 & 88.89 & \underline{96.97} & 92.75 \\
    GeistBERT & \textbf{93.33} & 87.50 & \underline{90.32} & 100 & 100 & 100 & 88.57 & 93.94 & 91.18 \\
    GeBERTa & \underline{84.38} & 84.38 & 84.38 & 100 & 100 & 100 & \textbf{96.97} & \underline{96.97} & \textbf{96.97} \\
    \midrule
    & \multicolumn{3}{c}{\bfseries Emergency} & 
    \multicolumn{3}{c}{\multirow{2}{*}{\bfseries Traumatology}} & 
    \multicolumn{3}{c}{\multirow{2}{*}{\bfseries Anesthesiology}} \\
    & \multicolumn{3}{c}{\bfseries Medicine} & & & \\
    \cmidrule(lr){2-4} \cmidrule(lr){5-7} \cmidrule(lr){8-10}
    \ChristBERT & 100 & 100 & 100 & 100 & 100 & 100 & 100 & 100 & 100 \\
    \ChristBERT\textsubscript{scratch} & 100 & 100 & 100 & 100 & 100 & 100 & 100 & 100 & 100 \\
    \ChristBERT\textsubscript{BPE} & 100 & 100 & 100 & 100 & 100 & 100 & 100 & 100 & 100 \\
    medBERT.de & 100 & 100 & 100 & 100 & 100 & 100 & 100 & 100 & 100 \\
    BioGottBERT & 100 & 100 & 100 & 100 & 100 & 100 & 100 & 100 & 100 \\
    GeistBERT & 100 & 100 & 100 & 100 & 100 & 100 & 100 & 100 & 100 \\
    GeBERTa & 100 & 100 & 100 & 100 & 100 & 100 & 100 & 100 & 100 \\
    \bottomrule
\end{tabular}
  
  \caption[Overview of per class precision, recall and \ff{} scores achieved on
  the JSynCC dataset]{Overview of per class precision (Prec.), recall (Rec.)
  and \ff{} scores achieved on the JSynCC dataset All results are shown in
  percent and assess each model's best fine-tuned performance on the test set.
  The best model was selected out of 28 runs based on its validation set
  performance. Best score in bold and second best underlined.}
  \label{tab:jsyncc}
\end{table}


\end{document}